\documentclass[letterpaper]{article} 
\usepackage{aaai2026}  
\usepackage{times}  
\usepackage{helvet}  
\usepackage{courier}  
\usepackage[hyphens]{url}  
\usepackage{graphicx} 
\usepackage{natbib}  
\usepackage{caption} 
\usepackage{appendix}
\usepackage{algorithm}
\usepackage{algorithmic}
\usepackage{multirow}
\usepackage{makecell}
\usepackage{amsmath} 
\usepackage{amssymb} 
\usepackage{lineno}

\usepackage{newfloat}
\usepackage{listings}
\DeclareCaptionStyle{ruled}{labelfont=normalfont,labelsep=colon,strut=off} 
\floatstyle{ruled}
\newfloat{listing}{tb}{lst}{}
\floatname{listing}{Listing}
\title{ Time Series Forecasting via Direct Per-Step \\ Probability Distribution Modeling}
\author{
    Linghao Kong,
    Xiaopeng Hong\thanks{Corresponding author}
}
\affiliations{
    The Faculty of Computing, Harbin Institute of Technology, Harbin, China\\

    leonardokong486@gmail.com, hongxiaopeng@ieee.org
}

\usepackage{bibentry}

\begin{document}

\maketitle

\begin{abstract}
Deep neural network-based time series prediction models have recently demonstrated superior capabilities in capturing complex temporal dependencies.
However, it is challenging for these models to account for uncertainty associated with their predictions, because they directly output scalar values at each time step.
To address such a challenge, we propose a novel model named \textbf{inter}leaved dual-branch \textbf{P}robability \textbf{D}istribution \textbf{N}etwork (\textbf{interPDN}), which directly constructs discrete probability distributions per step instead of a scalar.
The regression output at each time step is derived by computing the expectation of the predictive distribution on a predefined support set.
To mitigate prediction anomalies, a dual-branch architecture is introduced with interleaved support sets, augmented by coarse temporal-scale branches for long-term trend forecasting.
Outputs from another branch are treated as auxiliary signals to impose self-supervised consistency constraints on the current branch's prediction.
Extensive experiments on multiple real-world datasets demonstrate the superior performance of interPDN.
\end{abstract}

\begin{links}
    \link{Code}{https://github.com/leonardokong486/interPDN}
\end{links}


\section{Introduction}

As a highly challenging task in the field of time series, Time Series Forecasting (TSF) has attracted significant and growing attention in recent years.
TSF requires predicting data points over future horizons based on historical observations, while demonstrating strong demand across diverse domains such as finance, energy, healthcare, and transportation.
Within the mathematical frameworks employed in TSF tasks, neural network-based prediction models currently represent the most prevalent approach \cite{wang2024deep}.

\begin{figure}[t]
\centering
\includegraphics[width=0.9\columnwidth]{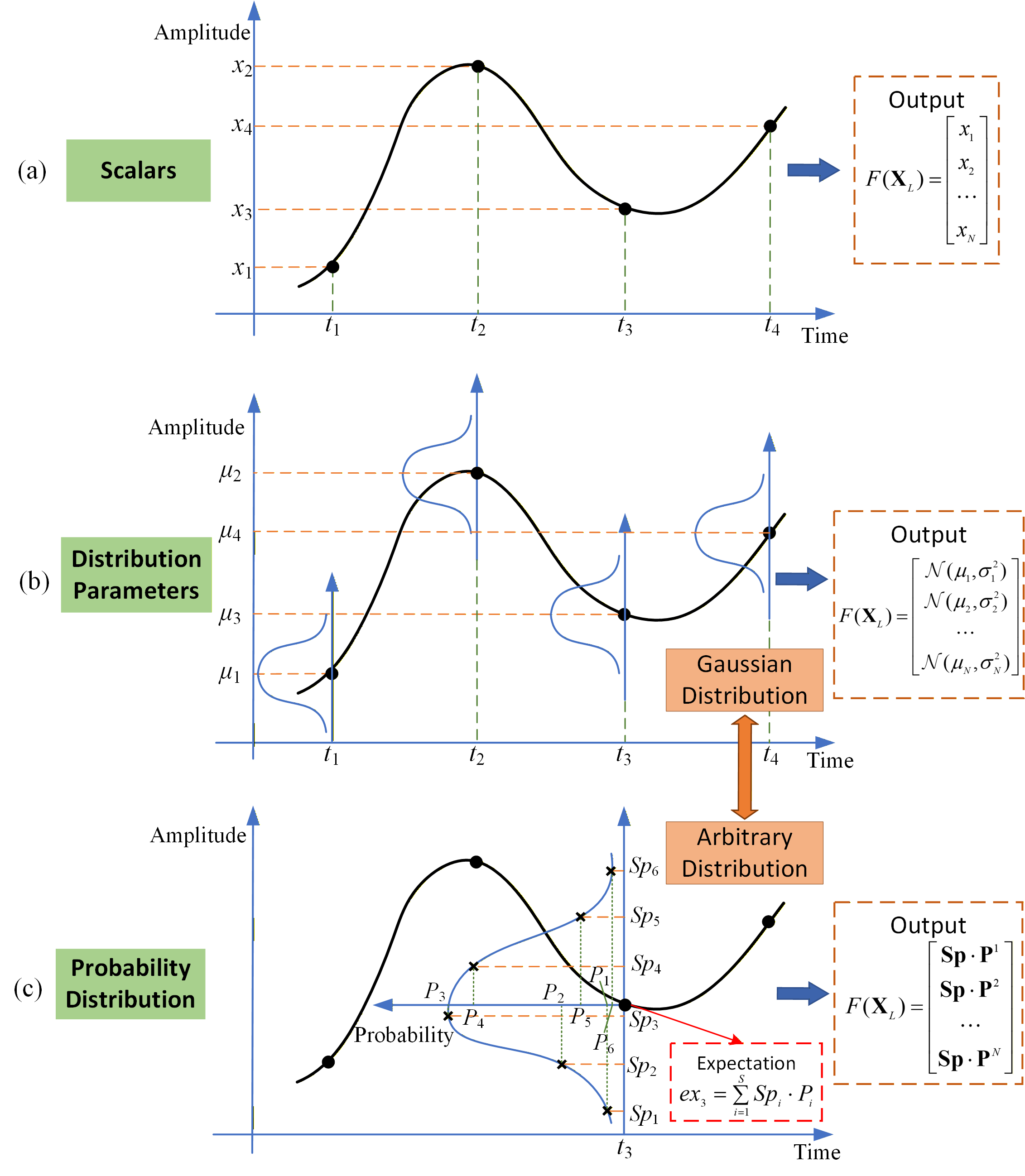} 
\caption{Different ways of per-step time series modeling: (a) Scalar estimation; (b) Probability distribution parameter prediction; (c) Expectation forecasting via discrete probability distribution}
\label{fig1}
\end{figure}

While most TSF models directly predict a sequence with scalars as demonstrated in Figure \ref{fig1}(a), AdaPTS \cite{benechehab2025adapts} outputs the mean and variance of the Gaussian distribution at each time step.
LangTime \cite{niu2025langtime} also estimates per-step probability distribution to compute Kullback-Leibler divergence from true distribution, which serves as the reward function for reinforcement learning policy.
These models output probability distribution parameters at each time step, providing users with interpretable confidence intervals, as shown in Figure \ref{fig1}(b).

Inspired by a crowd counting research \cite{lin2025semi} in Computer Vision (CV), we recast TSF as a direct probabilistic density distribution modeling problem.
The backbone does not output a scalar, instead it produces a discrete probability distribution over a predefined support set at each time step.
This approach provides advantages for capturing the inherent uncertainty in time series, while avoiding the prior assumption of Gaussian or other distributional forms required by models like AdaPTS.
Taking the expectation over the resulting probability distribution yields the prediction for each step, as illustrated in Figure \ref{fig1}(c).

Discrete probability distribution modeling requires quantization of output values, which can potentially introduce quantization errors and lead to misclassification at the bin boundary \cite{sun2025hierarchical}.
Therefore, we propose a dual-branch structure corresponding to interleaved support sets.
On one hand, it effectively restricts quantization errors and prevents forecasting anomalies with single branch.
On the other hand, the output additional signals enable self-supervised learning.
Since the model size doubles while the dataset remains unchanged, the constraints of dual-branch consistency losses are necessary to avoid training issues.

To steer the model towards accurate long-term trend forecasting without overfitting to local details, the interleaved dual-branch architecture is replicated at coarse time scale.
The coarse-scale output also serves as an additional self-supervised signal to suppress overfitting caused by doubled model parameters on the original small-scale dataset.
Consequently, a consistency loss is applied between the downsampled fine-grained outputs and the coarse-grained outputs.

We summarize our main contributions as follows:

\begin{itemize}
\item We transform the TSF task into a direct probability density distribution modeling problem.
At each time step, we output a discrete probability distribution, whose expected value serves as the final output.
\item To effectively mitigate the quantization error inherent in density distribution modeling, we propose an innovative interleaved dual-branch architecture coupled with corresponding self-supervised constraint.
\item Since the model struggles to discern trend evolution in long-term TSF under the normal time scale, we further introduce self-supervised constraints at a coarser scale.
\item The proposed interPDN has consistently outperformed state-of-the-art (SOTA) methods in multiple standardized real-world benchmark datasets.
\end{itemize}

\section{Related Work}

Statistical models used to lay foundation for traditional TSF models, such as Vector Autoregression (VAR) \cite{sims1980macroeconomics} and ARIMA \cite{box2013box}.
With the rapid rise of deep neural networks, TSF has drawn on classic models in the field of Natural Language Processing (NLP) and achieved more accurate predictions based on Recurrent Neural Networks (RNNs) \cite{salinas2020deepar} and Long-Short-Term Memory networks (LSTMs) \cite{rangapuram2018deep}.

Given that conventional TSF models struggle to capture long-term dependencies in sequences and handle non-stationary signals \cite{wang2024deep}, Transformer-based prediction models have gained increasing popularity.
Informer \cite{zhou2021informer} introduces the ProbSparse self-attention mechanism and attention distilling, which significantly reduces the computational complexity and memory footprint of Transformer architecture.
Autoformer \cite{wu2021autoformer} specifically targets the periodicity in time series and proposes computing attention between phase-correlated positions across different periods.
Unlike most channel-independent Transformer-based models, iTransformer \cite{liu2023itransformer} embeds a whole channel into a token to allow modeling variate-wise dependencies.
Subsequent research focuses mainly on token organization \cite{lu2024context}, design of attention layers \cite{lu2025linear}, and formulation of key-value pairs \cite{xue2023card} within Transformer models.

The aforementioned Transformer-based TSF models are typically small-scale, primarily due to the limited volume of time series datasets, making it challenging to train robust large models with strong generalization capabilities \cite{zhang2024large}.
Time-LLM \cite{jin2023time} reprograms time patches into text tokens and employs a frozen LLM as backbone for prediction of time series.
Frozen large image inpainting models also exhibit strong zero-shot performance in TSF tasks \cite{chen2024visionts}.
However, transferring pretrained models across domains introduces persistent challenges in explainability.

In contrast, compacting models with simple backbones demonstrates superior performance in TSF tasks, such as Multilayer Perceptrons (MLPs) and Convolutional Neural Networks (CNNs).
DLinear \cite{zeng2023transformers} employs a linear-only architecture to separately model trend and periodicity components in time series.
HDMixer \cite{huang2024hdmixer} introduces an adaptive-length patching mechanism that significantly enhances the forecasting capability of MLP architectures.
By breaking 1D time series down to 2D images based on periodicity, CNNs effectively expand their receptive fields beyond sequential constraints to capture richer temporal dependencies \cite{wu2022timesnet}.
Our model also builds upon a similarly lightweight backbone architecture, while drawing inspiration from existing multi-branch and multi-scale methodologies.

Unlike our interleaved branches, prior TSF models leveraged multi-branch architectures to independently process markedly distinct input features.
For instance, after decomposing the input sequence into seasonal and trend components using the moving average technique, separate prediction streams can be employed to forecast each component based on their unique statistical characteristics \cite{ma2024ts3net,lin2024cyclenet}.
In terms of learning in frequency-domain, Koopa \cite{liu2023koopa} decomposes time series into time-varying and time-invariant branches, which are modeled separately through local and global Koopman operators.
Time-VLM \cite{zhong2025time} transforms raw time series into images while deriving corresponding text descriptions of input, subsequently feeding temporal, visual, and textual representations into three dedicated channels.

In contrast to our practice of utilizing coarse-grained sequences as self-supervised constraint signals, other TSF models typically apply multi-scale techniques during patching stage \cite{chen2024pathformer,cho2025comres,tang2025unlocking, han2025retrieval}.
Unlike parallel multi-scale architectures, TTM \cite{ekambaram2024tiny} sequentially incorporates multi-scale patches at distinct hierarchical stages of the model.
Pyraformer \cite{liu2022pyraformer} organizes multi-scale time series into a pyramid architecture, performing sparse attention in adjacent steps and layers.
Some models also focus on mixing of multi-scale branches, such as AMD \cite{hu2025adaptive} and TimeMixer \cite{wang2024timemixer}.

\section{Proposed Method}
In multivariate TSF, given a historically observed sequence $\mathbf{X}_{L}=\left\{\mathbf{x}_{t-L+1},\cdots\cdots,\mathbf{x}_{t-1},\mathbf{x}_{t}\right\}$ with a length of $L$, the goal is to predict the target values over the next $T$ time steps $\mathbf{X}_{T}=\left\{\mathbf{x}_{t+1},\mathbf{x}_{t+2},\cdots\cdots,\mathbf{x}_{t+T}\right\}$. Herein, $\mathbf{x}_{t} \in \mathbb{R}^{1 \times C}$ denotes the multivariate data at time step $t$, and $C$ represents the number of channels.

\begin{figure*}[t]
\centering
\includegraphics[width=0.8\textwidth]{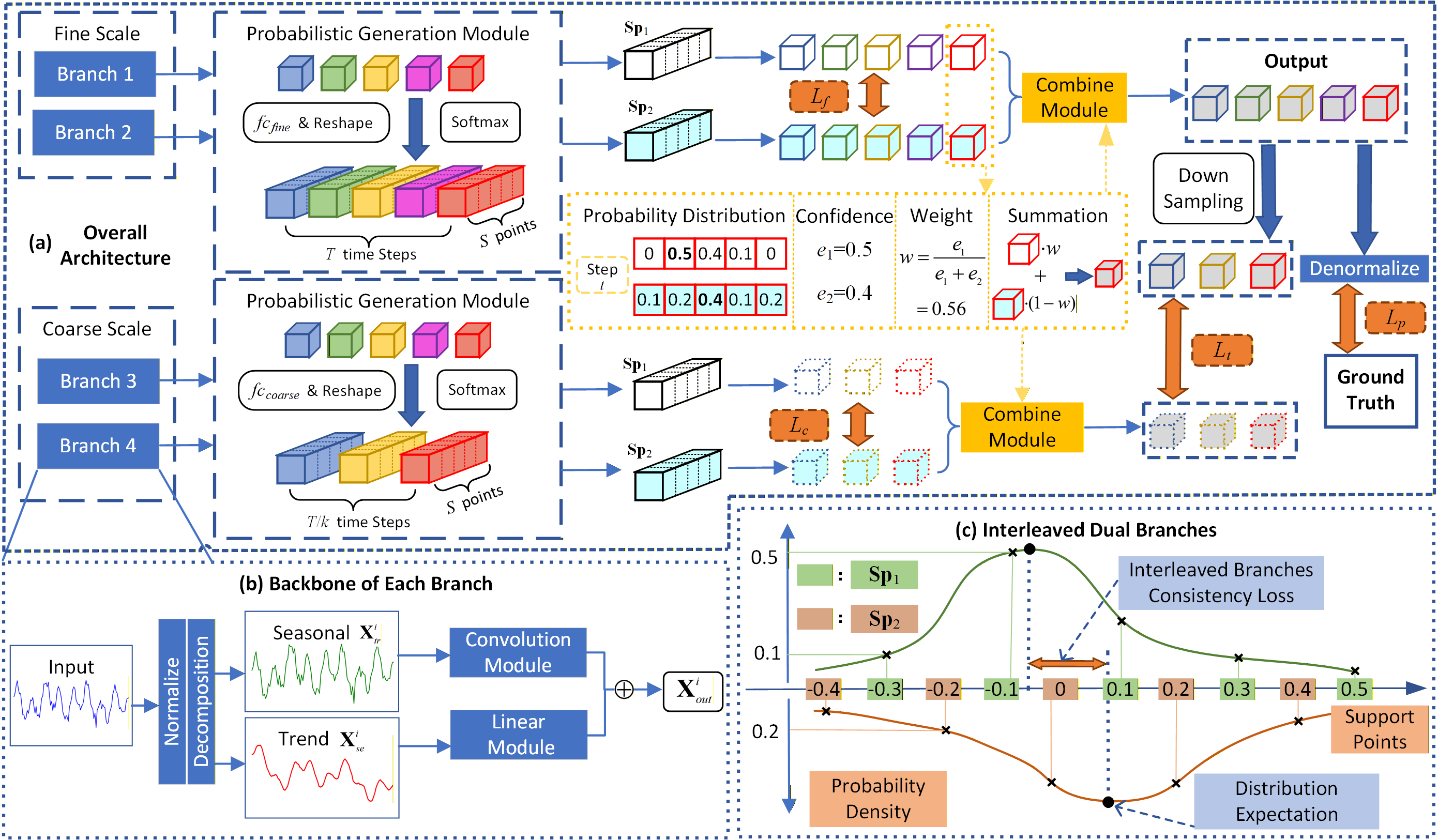}
\caption{(a) Overall model architecture for direct per-step probability distribution forecasting ; (b) Backbone structure of each branch; (c) Dual-branch design corresponding to interleaved support sets.}
\label{fig2}
\end{figure*}

The overall architecture of interPDN is depicted in Figure \ref{fig2}(a).
Within each branch, the backbone decomposes the raw input data into seasonal and trend components, which are then processed separately by convolutional and linear modules illustrated in Figure \ref{fig2}(b).
To output a per-step probability distribution, probabilistic generation module is introduced.
At each distinct time scale, the dual branches produce distributions associated with interleaved support sets, which are shown in Figure \ref{fig2}(c), and their outputs are mixed up by combine modules.
The dual branches are replicated at the coarse time scale, so that their outputs guide the model to focus on long-term trends.
Since outputs from the additional branch are treated as self-supervised signals, inter-scale and cross-scale consistency losses can constrain the model to prevent prediction deviations.

\subsection{Backbone of Each Branch}
To address variable lags and cross-channel interference \cite{zhao2024rethinking,yang2024vcformer} in multivariate TSF, the input multivariate time series $\mathbf{X}_{L}$ is split into $C$ single-channel sequences $\mathbf{X}_{L}^{i} \in \mathbb{R}^{L \times 1}$, following the channel-independent process \cite{nie2022time}.
$\mathbf{X}_{L}^{i}$ is normalized instance-wise through RevIN method \cite{kim2021reversible} to adapt to distribution shift.
Using exponential moving average method, trend component $\mathbf{X}_{tr}^{i}$ is extracted from the single-channel sequence, while the residual becomes seasonal component $\mathbf{X}_{se}^{i}$.

The seasonal processing stream starts with patching, which is a classic method in TSF for handling longer lookback windows \cite{nie2022time}.
Input $\mathbf{X}_{se}^{i}$ thus turns to $\mathbf{X}_{se,p}^{i} \in \mathbb{R}^{N \times P}$ with the stride length $S_t$ and patch length $P$, and the number of patches $N$ can be calculated by $N=\left\lfloor\frac{(L-P)}{S_t}\right\rfloor+2$. 
A linear transformation followed by an activation function is applied to the patch count dimension of $\mathbf{X}_{se,p}^{i}$.
Subsequently, 1D convolution is employed along this dimension to aggregate information from adjacent patches.
The outputs of the linear block and the convolutional block are then integrated via a ResNet structure \cite{he2016deep} and fed into an MLP decoder to obtain $\mathbf{X}_{se,out}^{i} \in \mathbb{R}^{1 \times T}$.


Another stream for trend processing builds upon two linear blocks without any activation layers, thus transforming $\mathbf{X}_{tr}^{i}$ into $\mathbf{X}_{tr,out}^{i}\in \mathbb{R}^{1 \times T}$. 
Each linear block requires the application of average pooling and layer normalization subsequent to its fully connected layer.
The results of seasonal and trend streams are concatenated along the last dimension to obtain final output $\mathbf{X}_{out}^{i} \in \mathbb{R}^{1 \times 2T}$ of each backbone branch:
\begin{equation}
\mathbf{X}_{out}^{i} = \mathbf{X}_{tr,out}^{i} \oplus \mathbf{X}_{se,out}^{i}.
\end{equation}


\subsection{Probabilistic Generation Module}
\label{PGM}

In conventional TSF models that directly produce scalar predictions, an output projection head is typically used to reduce the dimensionality of $\mathbf{X}_{out}$ and obtain the prediction target $\hat{\mathbf{X}}_{T}$.
However, such output heads fail to account for the uncertainty inherent in TSF, particularly by producing low-confidence predictions at challenging steps in the time sequences.
In interPDN, we introduce a probabilistic generation module that outputs per-step discrete probability distribution rather than a scalar.
Critically, we make no assumptions about the form of this distribution (e.g., Gaussian, mixture Gaussian, or multinomial or other parametric families).
Furthermore, the module explicitly generates the full probability distribution directly, rather than inferring it indirectly via distributional parameters.

A fully connected layer $fc_{fine}$ is employed to expand the dimension of the branch backbone output $\mathbf{X}_{out}^{i}$ from $2T$ to $S \times T$, where $S$ represents the number of elements in the predefined support set. Subsequently, the output is reshaped:
\begin{equation}
\mathbf{X}_{f}^{i} = \mathrm{Reshape} \left( fc_{fine} \left( \mathbf{X}_{out}^{i} \right) \right),
\end{equation}
thus forming $\mathbf{X}_{f}^{i} \in \mathbb{R}^{T \times S}$.

Given that normalized time-series data tends to follow a normal distribution, applying uniformly spaced partition points to the support set is inadvisable for TSF tasks.
It is preferable to partition the sub-intervals of the support set into non-uniform, equal-probability intervals.
This strategy maximizes the proximity of data points to support points while minimizing their occurrences in interval centers, thus mitigating categorical errors.
The Cumulative Distribution Function (CDF) of the normal distribution and its inverse function can be utilized to solve for breakpoints that divide the restricted interval $\left[-B, B\right]$ into $(S-1)$ sub-intervals with equal probability.
By arranging these breakpoints in sequence along with the two boundary points $\pm B$, and calculating mean value of each pair of adjacent points, the support set vector $\mathbf{Sp}_{1} \in \mathbb{R}^{S \times 1}$ can be obtained.
At each time step, the support set remains identical.

In the case of a single branch with the output denoted as $\mathbf{X}_{f,1}^{i}$, the final prediction $\mathbf{X}_{sp,1}^{i}$ is obtained by computing the expectation of the output probability distribution over this predefined support set:
\begin{equation}
    \mathbf{X}_{sp,1}^{i} = \langle \mathrm{Softmax}\bigl( \mathbf{X}_{f,1}^{i} \bigr) , \mathbf{Sp}_{1}\rangle,
\label{cls1}
\end{equation}
where $\mathbf{X}_{sp,1}^{i} \in \mathbb{R}^{T \times 1}$ and $\langle \rangle$ denotes the vector dot product.

\subsection{Interleaved Dual Branches}
\label{IDB}

Compared with the probability distribution on a single set, the interleaved design with dual branches is more conducive in improving the robustness of the model.
When severe outliers occur in the predictions of one branch, the other branch can be relied upon for correction whose outputs also become additional self-supervised signals.
Moreover, when the ground truth at a certain time step falls in the middle of two support points of one branch, quantization error may occur, which is known as boundary effect \cite{sun2025hierarchical}.
Introducing another branch with interleaved support set thus mitigates the misclassification near boundaries of adjacent bins.
To create another set with interleaved support points, the mean value of adjacent points in $\mathbf{Sp}_{1}$ is calculated, and a predefined boundary point is simultaneously supplemented to form another support set vector $\mathbf{Sp}_{2}$.

On the normal time scale, $\mathbf{X}_{out,1}^{i}$ and $\mathbf{X}_{out,2}^{i}$ are computed using two branches with independent parameters.
Then $\mathbf{X}_{f,1}^{i}$ and $\mathbf{X}_{f,2}^{i}$ are derived by mapping through two fully connected layers ($fc_{fine,1}$ and $fc_{fine,2}$) that also have independent parameters.
Probability distributions on the second support set are generated through softmax in the last dimension of $\mathbf{X}_{f,2}^{i}$, and the expectation $\mathbf{X}_{sp,2}^{i}$ can be calculated as Equation \ref{cls1}.

Unlike existing approaches that aggregate multi-branch outputs through summation \cite{zhang2022first}, we come up with a method based on maximum prediction confidence.
First, it is necessary to calculate the maximum probability value $\mathbf{e}_{1}, \mathbf{e}_{2} \in \mathbb{R}^{T \times 1}$ across all support points in the prediction for each time step: 
\begin{equation}
\begin{split}
\mathbf{e}_{1} = \max_{1 \leq s \leq S} \Bigl( & \mathrm{Softmax}\bigl(\mathbf{X}_{f,1}^{i}\bigr) \Bigr)_{t,s}, \\
& t=1,2,\cdots\cdots, T,
\end{split}
\label{e1}
\end{equation}
\begin{equation}
\begin{split}
\mathbf{e}_{2} = \max_{1 \leq s \leq S} \Bigl( & \mathrm{Softmax}\bigl(\mathbf{X}_{f,2}^{i}\bigr) \Bigr)_{t,s}, \\
& t=1,2,\cdots\cdots, T.
\end{split}
\label{e2}
\end{equation}
Second, the weight coefficient $\mathbf{w}$ is calculated based on the maximum confidence of the two branches:
\begin{equation}
\mathbf{w} = \frac{\mathbf{e}_1}{\mathbf{e}_1 + \mathbf{e}_{2}}.
\label{w}
\end{equation}
Third, a weighted sum is performed on the resulting expectations of the two interleaved branches:
\begin{equation}
\mathbf{X}_{sp,f}^{i} = \mathbf{w} \odot \mathbf{X}_{sp,1}^{i} + (1 - \mathbf{w}) \odot \mathbf{X}_{sp,2}^{i},
\label{clsfine}
\end{equation}
where $\odot$ denotes the Hadamard product.
By applying the RevIN denormalization method to $\mathbf{X}_{sp,f}^{i} \in \mathbb{R}^{T \times 1}$ and reassembling the separated channels, the final output $\hat{\mathbf{X}}_T \in \mathbb{R}^{T \times C}$ of the model can be obtained.

\subsection{Bi-scale Temporal Branches}

To further constrain the fusion output on the normal (fine) time scale, we replicate the dual-branch design mentioned in Section \ref{PGM} at the coarse scale to generate auxiliary self-supervised sequences in lower resolution.
In this way, the overall model contains a total of four backbone branches with the same architecture but non-shared parameters as shown in the left part of Figure \ref{fig2}(a): two backbone branches used at coarse time scale while another two backbone branches used at fine time scale.

We adjust the projection layer of the probability generation module mentioned in Section \ref{PGM}, thus altering the output dimension of this module:
\begin{equation}
\mathbf{X}_{c}^{i} = \mathrm{Reshape} \left( fc_{coarse} \left( \mathbf{X}_{out}^{i} \right) \right).
\end{equation}
The output dimension of the fully connected layer $fc_{coarse}$ is no longer $T \times S$ but $\frac{T}{k} \times S$ instead, thus generating $\mathbf{X}_{c}^{i} \in \mathbb{R}^{\frac{T}{k} \times S}$. $k$ is a positive integer downsampling factor greater than 1.
$T$ is divisible by $k$, otherwise padding must be applied to the dimension of sequence length.
All four fully connected layers ($fc_{coarse,1}$, $fc_{coarse,2}$, $fc_{fine,1}$, and $fc_{fine,2}$) employed in the probability generation module do not share any parameters.

The interleaved support sets are the same as $\mathbf{Sp}_{1}$ and $\mathbf{Sp}_{2}$, so the subsequent expectation calculation process is identical to that in Equation \ref{cls1}.
The expected value calculated at the coarse scale is denoted as $\mathbf{X}_{sp,3}^{i} , \mathbf{X}_{sp,4}^{i} \in \mathbb{R}^{\frac{T}{k} \times 1}$.
The calculation process of the weight coefficient $\mathbf{w}$ for fusing the forecast results of the two branches is also the same as that in Equations \ref{e1} to \ref{w}.
Final result of the coarse time-scale branch $\mathbf{X}_{sp,c}^{i} \in \mathbb{R}^{\frac{T}{k} \times 1}$ is generated by the following formula:
\begin{equation}
\mathbf{X}_{sp,c}^{i} = \mathbf{w} \odot \mathbf{X}_{sp,3}^{i} + (1 - \mathbf{w}) \odot \mathbf{X}_{sp,4}^{i}.
\label{clscoase}
\end{equation}
It should be noted that $\mathbf{X}_{sp,c}^{i}$ serves solely as a self-supervised signal that constrains the output of the fine-scale branches, and it is not combined with $\hat{\mathbf{X}}_T$.

\subsection{Loss Function}

Previous TSF models often employ the Mean Square Error (MSE) loss for backpropagation. However, when the prediction length is long, using only the MSE loss as the objective function makes it difficult for the model to capture the continuity and pattern correlation between consecutive time steps.
In severe cases, this can lead to direction errors or pattern distortion \cite{wang2024fredf, kudrat2025patch}.

InterPDN refers to the loss function adopted by the xPatch prediction model \cite{stitsyuk2025xpatch}. 
xPatch assigns different weights to errors at various time steps.
In particular, the weights of prediction errors on steps closer to the lookback window are higher, while the weights of errors at more distant time steps decay.
The loss between the predictions $\hat{\mathbf{X}}_{T}$ and the ground truth $\mathbf{X}_{T}$ is written as $L_{p}$, which can be calculated by:
\begin{equation}
L_{p}\left( \mathbf{X}_{T}, \hat{\mathbf{X}}_{T} \right) 
= \frac{1}{T \cdot C} \sum_{i=1}^{T} \theta(i) 
\left\| \mathbf{x}_{t+i} - \hat{\mathbf{x}}_{t+i} \right\|_{1},
\end{equation}
where $\theta(i)$ is an error decay function based on the arctangent function.

Since the proposed multi-branch model possesses four times the parameters of the original backbone while the dataset remains unchanged, it is prone to issues such as overfitting and degradation in generalization performance.
Hence, similar to self-supervised learning, outputs from another branch can serve as additional supervisory signals to impose consistency constraints on the output of the present branch.
Moreover, if results of interleaved set branches are not constrained, it is highly likely that one branch would exhibit high-confidence prediction while the other would perform severe quantization errors in distribution forecasting.
Consequently, it can be difficult to achieve the effects of output complementarity or resolve misclassification at bin boundary as mentioned in Section \ref{IDB}.

We employ MSE to impose inter-scale consistency constraint loss.
We denote by $L_{f}$ the consistency loss between outputs $\mathbf{X}_{sp, 1}$ and $\mathbf{X}_{sp, 2}$ of the fine-scale dual branches, while by $L_{c}$ the consistency loss between outputs $\mathbf{X}_{sp, 3}$ and $\mathbf{X}_{sp, 4}$ of the coarse-scale dual branches, respectively:
\begin{equation}
L_{f}\left( \mathbf{X}_{sp, 1}, \mathbf{X}_{sp, 2} \right) = \frac{1}{T \cdot C} \sum_{i=1}^{C} \left\| \mathbf{X}_{sp, 1}^{i} - \mathbf{X}_{sp, 2}^{i} \right\|_{2}^{2} ,
\end{equation}
\begin{equation}
L_{c}\left( \mathbf{X}_{sp, 3}, \mathbf{X}_{sp, 4} \right) = \frac{k}{T \cdot C} \sum_{i=1}^{C} \left\| \mathbf{X}_{sp, 3}^{i} - \mathbf{X}_{sp, 4}^{i} \right\|_{2}^{2}.
\end{equation}

Multi-scale prediction branches are at risk of learning contradictory patterns from different time resolutions \cite{zhang2025waveletmixer}.
To enable the coarse-grained branch to guide the fine-scale branch in terms of long-term evolutionary trends, it is necessary to introduce an inter-scale consistency loss function.
Due to the discrepancy in the number of time steps between $\mathbf{X}_{sp,f}^{i}$ and $\mathbf{X}_{sp,c}^{i}$, it is necessary to downsample $\mathbf{X}_{sp,f}^{i}$ along the time dimension to form $\mathbf{X}_{d,f}^{i}$:
\begin{equation} 
\mathbf{X}_{d,f}^{i} 
= \mathrm{AvgPool}\bigl( 
    \mathbf{X}_{sp,f}^{i}, 
    \mathrm{kernel}=k 
\bigr),
\end{equation}
where both kernel size and stride length in the average pooling operation are set to $k$.
We compute the cross-scale consistency loss $L_{t}$ between $\mathbf{X}_{d,f}^{i}$ and $\mathbf{X}_{sp,c}^{i}$ still using MSE:
\begin{equation}
L_{t} \left( \mathbf{X}_{sp,c}, \mathbf{X}_{d,f} \right) = \frac{k}{T \cdot C} \sum_{i=1}^{C} \bigl\| \mathbf{X}_{sp,c}^{i} - \mathbf{X}_{d,f}^{i} \bigr\|_{2}^{2}.
\end{equation}

The total loss $L_{total}$ involved in backpropagation consists of the following four components: the loss between the predicted value and the true value, two consistency losses of interleaved set branches, and the cross-scale consistency loss.
The calculation formula of $L_{total}$ is as follows:
\begin{equation}
L_{total} = L_{p} + \alpha \cdot L_{f} + \beta \cdot L_{c} + \gamma \cdot L_{t},
\end{equation}
where $\alpha, \beta, \gamma \geq 0$ are the weight coefficients used to balance various kinds of losses.

\section{Experiments}

\begin{table*}[!t]
 \centering 
  \small 
  \resizebox{\textwidth}{!}{ 
  \renewcommand{\arraystretch}{1.1}
\begin{tabular}{cc|cc|cc|cc|cc|cc|cc|cc|cc|cc|cc}
    \hline
    \multicolumn{2}{c|}{Models} & 
    \multicolumn{2}{c|}{\makecell{\textbf{interPDN} \\ (\textbf{Ours})}} & 
    \multicolumn{2}{c|}{\makecell{xPatch \\ (2025)}} & 
    \multicolumn{2}{c|}{\makecell{RAFT \\ (2025)}} & 
    \multicolumn{2}{c|}{\makecell{AMD \\ (2025)}} & 
    \multicolumn{2}{c|}{\makecell{MOMENT \\ (2024)}} & 
    \multicolumn{2}{c|}{\makecell{TimeMixer \\ (2024)}} & 
    \multicolumn{2}{c|}{\makecell{iTransformer \\ (2024)}} & 
    \multicolumn{2}{c|}{\makecell{TimesNet \\ (2023)}} & 
    \multicolumn{2}{c|}{\makecell{PatchTST \\ (2023)}} & 
    \multicolumn{2}{c}{\makecell{DLinear \\ (2023)}} \\
    \hline
\multicolumn{2}{c|}{Metric}                             & MSE              & MAE             & MSE            & MAE               & MSE             & MAE            & MSE            & MAE            & MSE              & MAE             & MSE               & MAE               & MSE                 & MAE                & MSE               & MAE              & MSE                & MAE             & MSE              & MAE             \\ \hline
\multicolumn{1}{c|}{\multirow{5}{*}{ETTh1}}       & 96  & \textbf{0.338}   & \textbf{0.374}  & {\underline{0.355}}    & {\underline{0.382}}       & 0.367           & 0.397          & 0.369          & 0.397          & 0.387            & 0.410           & 0.361             & 0.390             & 0.386               & 0.405              & 0.384             & 0.402            & 0.370              & 0.399           & 0.384            & 0.405           \\
\multicolumn{1}{c|}{}                             & 192 & \textbf{0.362}   & \textbf{0.391}  & {\underline{0.381}}    & {\underline{0.396}}       & 0.411           & 0.427          & 0.401          & 0.417          & 0.410            & 0.426           & 0.409             & 0.414             & 0.441               & 0.436              & 0.436             & 0.429            & 0.413              & 0.421           & 0.405            & 0.413           \\
\multicolumn{1}{c|}{}                             & 336 & \textbf{0.378}   & \textbf{0.406}  & {\underline{0.398}}    & {\underline{0.413}}       & 0.436           & 0.442          & 0.418          & 0.427          & 0.422            & 0.437           & 0.430             & 0.429             & 0.487               & 0.458              & 0.491             & 0.469            & 0.422              & 0.436           & 0.447            & 0.448           \\
\multicolumn{1}{c|}{}                             & 720 & \textbf{0.432}   & \textbf{0.451}  & 0.446          & 0.456             & 0.467           & 0.478          & {\underline{0.439}}    & {\underline{0.454}}    & 0.454            & 0.472           & 0.445             & 0.460             & 0.503               & 0.491              & 0.521             & 0.500            & 0.447              & 0.468           & 0.471            & 0.487           \\ \cline{2-22} 
\multicolumn{1}{c|}{}                             & avg & \textbf{0.378}   & \textbf{0.406}  & {\underline{0.395}}    & {\underline{0.412}}       & 0.420           & 0.436          & 0.407          & 0.424          & 0.418            & 0.436           & 0.411             & 0.423             & 0.454               & 0.448              & 0.458             & 0.450            & 0.413              & 0.431           & 0.427            & 0.438           \\ \hline
\multicolumn{1}{c|}{\multirow{5}{*}{ETTh2}}       & 96  & \textbf{0.223}   & \textbf{0.297}  & {\underline{0.229}}    & {\underline{0.299}}       & 0.276           & 0.344          & 0.274          & 0.337          & 0.288            & 0.345           & 0.271             & 0.330             & 0.297               & 0.349              & 0.340             & 0.374            & 0.274              & 0.336           & 0.289            & 0.353           \\
\multicolumn{1}{c|}{}                             & 192 & \textbf{0.267}   & \textbf{0.328}  & {\underline{0.276}}    & {\underline{0.334}}       & 0.347           & 0.393          & 0.351          & 0.383          & 0.349            & 0.386           & 0.317             & 0.402             & 0.380               & 0.400              & 0.402             & 0.414            & 0.339              & 0.379           & 0.383            & 0.418           \\
\multicolumn{1}{c|}{}                             & 336 & \textbf{0.305}   & \textbf{0.360}  & {\underline{0.315}}    & {\underline{0.365}}       & 0.376           & 0.425          & 0.375          & 0.411          & 0.369            & 0.408           & 0.332             & 0.396             & 0.428               & 0.432              & 0.452             & 0.452            & 0.329              & 0.380           & 0.448            & 0.465           \\
\multicolumn{1}{c|}{}                             & 720 & {\underline{0.377}}      & {\underline{0.416}}     & 0.382          & 0.418             & 0.436           & 0.473          & 0.402          & 0.438          & 0.403            & 0.439           & \textbf{0.342}    & \textbf{0.408}    & 0.427               & 0.445              & 0.462             & 0.468            & 0.379              & 0.422           & 0.605            & 0.551           \\ \cline{2-22} 
\multicolumn{1}{c|}{}                             & avg & \textbf{0.293}   & \textbf{0.350}  & {\underline{0.301}}    & {\underline{0.354}}       & 0.359           & 0.409          & 0.351          & 0.392          & 0.352            & 0.395           & 0.316             & 0.384             & 0.383               & 0.407              & 0.414             & 0.427            & 0.330              & 0.379           & 0.431            & 0.447           \\ \hline
\multicolumn{1}{c|}{\multirow{5}{*}{ETTm1}}       & 96  & \textbf{0.277}   & \textbf{0.331}  & {\underline{0.280}}    & {\underline{0.333}}       & 0.302           & 0.349          & 0.284          & 0.339          & 0.293            & 0.349           & 0.291             & 0.340             & 0.334               & 0.368              & 0.338             & 0.375            & 0.290              & 0.342           & 0.301            & 0.345           \\
\multicolumn{1}{c|}{}                             & 192 & \textbf{0.310}   & \textbf{0.352}  & {\underline{0.318}}    & {\underline{0.356}}       & 0.329           & 0.367          & 0.322          & 0.362          & 0.326            & 0.368           & 0.327             & 0.365             & 0.377               & 0.391              & 0.374             & 0.387            & 0.332              & 0.369           & 0.336            & 0.366           \\
\multicolumn{1}{c|}{}                             & 336 & \textbf{0.348}   & \textbf{0.372}  & 0.357          & 0.380             & 0.355           & 0.383          & 0.361          & 0.383          & {\underline{0.352}}      & {\underline{0.379}}     & 0.360             & 0.381             & 0.426               & 0.420              & 0.410             & 0.411            & 0.366              & 0.392           & 0.372            & 0.389           \\
\multicolumn{1}{c|}{}                             & 720 & 0.411            & \textbf{0.404}  & 0.422          & 0.414             & {\underline{0.406}}     & {\underline{0.413}}    & 0.421          & 0.418          & \textbf{0.405}   & 0.416           & 0.415             & 0.417             & 0.491               & 0.459              & 0.478             & 0.450            & 0.416              & 0.420           & 0.427            & 0.423           \\ \cline{2-22} 
\multicolumn{1}{c|}{}                             & avg & \textbf{0.337}   & \textbf{0.365}  & {\underline{0.344}}    & {\underline{0.371}}       & 0.348           & 0.378          & 0.347          & 0.376          & {\underline{0.344}}      & 0.379           & 0.348             & 0.376             & 0.407               & 0.410              & 0.400             & 0.406            & 0.351              & 0.381           & 0.359            & 0.381           \\ \hline
\multicolumn{1}{c|}{\multirow{5}{*}{ETTm2}}       & 96  & \textbf{0.149}   & \textbf{0.237}  & {\underline{0.151}}    & {\underline{0.240}}       & 0.164           & 0.256          & 0.167          & 0.258          & 0.170            & 0.260           & 0.164             & 0.254             & 0.180               & 0.264              & 0.187             & 0.267            & 0.165              & 0.255           & 0.172            & 0.267           \\
\multicolumn{1}{c|}{}                             & 192 & \textbf{0.208}   & \textbf{0.277}  & {\underline{0.211}}    & {\underline{0.281}}       & 0.219           & 0.296          & 0.221          & 0.295          & 0.227            & 0.297           & 0.223             & 0.295             & 0.250               & 0.309              & 0.249             & 0.309            & 0.220              & 0.292           & 0.237            & 0.314           \\
\multicolumn{1}{c|}{}                             & 336 & \textbf{0.263}   & \textbf{0.314}  & {\underline{0.267}}    & {\underline{0.318}}       & 0.275           & 0.336          & 0.270          & 0.327          & 0.275            & 0.328           & 0.279             & 0.330             & 0.311               & 0.348              & 0.321             & 0.351            & 0.274              & 0.329           & 0.278            & 0.338           \\
\multicolumn{1}{c|}{}                             & 720 & \textbf{0.336}   & \textbf{0.362}  & {\underline{0.343}}    & {\underline{0.368}}       & 0.359           & 0.392          & 0.356          & 0.382          & 0.363            & 0.387           & 0.359             & 0.383             & 0.412               & 0.407              & 0.408             & 0.403            & 0.362              & 0.385           & 0.420            & 0.437           \\ \cline{2-22} 
\multicolumn{1}{c|}{}                             & avg & \textbf{0.239}   & \textbf{0.298}  & {\underline{0.243}}    & {\underline{0.302}}       & 0.254           & 0.320          & 0.254          & 0.316          & 0.259            & 0.318           & 0.256             & 0.316             & 0.288               & 0.332              & 0.291             & 0.333            & 0.255              & 0.315           & 0.277            & 0.339           \\ \hline
\multicolumn{1}{c|}{\multirow{5}{*}{Weather}}     & 96  & \textbf{0.144}   & \textbf{0.180}  & 0.147          & {\underline{0.185}}       & 0.165           & 0.222          & {\underline{0.145}}    & 0.198          & 0.154            & 0.209           & 0.147             & 0.197             & 0.174               & 0.214              & 0.172             & 0.220            & 0.149              & 0.198           & 0.176            & 0.237           \\
\multicolumn{1}{c|}{}                             & 192 & \textbf{0.188}   & \textbf{0.223}  & 0.198          & {\underline{0.229}}       & 0.211           & 0.264          & 0.190          & 0.240          & 0.197            & 0.248           & {\underline{0.189}}       & 0.239             & 0.221               & 0.254              & 0.219             & 0.261            & 0.194              & 0.241           & 0.220            & 0.282           \\
\multicolumn{1}{c|}{}                             & 336 & \textbf{0.220}   & \textbf{0.257}  & {\underline{0.222}}    & {\underline{0.260}}       & 0.260           & 0.302          & 0.242          & 0.280          & 0.246            & 0.285           & 0.241             & 0.280             & 0.278               & 0.296              & 0.280             & 0.306            & 0.245              & 0.282           & 0.265            & 0.319           \\
\multicolumn{1}{c|}{}                             & 720 & \textbf{0.294}   & \textbf{0.310}  & {\underline{0.298}}    & {\underline{0.316}}       & 0.327           & 0.355          & 0.315          & 0.332          & 0.315            & 0.336           & 0.310             & 0.330             & 0.358               & 0.347              & 0.365             & 0.359            & 0.314              & 0.334           & 0.323            & 0.362           \\ \cline{2-22} 
\multicolumn{1}{c|}{}                             & avg & \textbf{0.212}   & \textbf{0.243}  & {\underline{0.216}}    & {\underline{0.248}}       & 0.241           & 0.286          & 0.223          & 0.263          & 0.228            & 0.270           & 0.222             & 0.262             & 0.258               & 0.278              & 0.259             & 0.287            & 0.226              & 0.264           & 0.246            & 0.300           \\ \hline
\multicolumn{1}{c|}{\multirow{5}{*}{Traffic}}     & 96  & {\underline{0.364}}      & \textbf{0.229}  & 0.367          & {\underline{0.237}}       & 0.378           & 0.273          & 0.366          & 0.259          & 0.391            & 0.282           & \textbf{0.360}    & 0.249             & 0.395               & 0.268              & 0.593             & 0.321            & \textbf{0.360}     & 0.249           & 0.410            & 0.282           \\
\multicolumn{1}{c|}{}                             & 192 & {\underline{0.377}}      & \textbf{0.239}  & 0.378          & {\underline{0.241}}       & 0.391           & 0.277          & 0.381          & 0.265          & 0.404            & 0.287           & \textbf{0.375}    & 0.250             & 0.417               & 0.276              & 0.617             & 0.336            & 0.379              & 0.256           & 0.423            & 0.287           \\
\multicolumn{1}{c|}{}                             & 336 & {\underline{0.389}}      & \textbf{0.240}  & 0.391          & {\underline{0.246}}       & 0.402           & 0.282          & 0.397          & 0.273          & 0.414            & 0.292           & \textbf{0.385}    & 0.270             & 0.433               & 0.283              & 0.629             & 0.336            & 0.392              & 0.264           & 0.436            & 0.296           \\
\multicolumn{1}{c|}{}                             & 720 & 0.433            & \textbf{0.269}  & 0.435          & {\underline{0.271}}       & 0.434           & 0.297          & \textbf{0.430} & 0.292          & 0.450            & 0.310           & \textbf{0.430}    & 0.281             & 0.467               & 0.302              & 0.640             & 0.350            & {\underline{0.432}}        & 0.286           & 0.466            & 0.315           \\ \cline{2-22} 
\multicolumn{1}{c|}{}                             & avg & {\underline{0.391}}      & \textbf{0.244}  & 0.393          & {\underline{0.249}}       & 0.401           & 0.282          & 0.394          & 0.272          & 0.415            & 0.293           & \textbf{0.388}    & 0.263             & 0.428               & 0.282              & 0.620             & 0.336            & {\underline{0.391}}        & 0.264           & 0.434            & 0.295           \\ \hline
\multicolumn{1}{c|}{\multirow{5}{*}{Electricity}} & 96  & \textbf{0.125}   & \textbf{0.215}  & {\underline{0.126}}    & {\underline{0.217}}       & 0.133           & 0.232          & 0.129          & 0.225          & 0.136            & 0.233           & 0.129             & 0.224             & 0.148               & 0.240              & 0.168             & 0.272            & 0.129              & 0.222           & 0.140            & 0.237           \\
\multicolumn{1}{c|}{}                             & 192 & \textbf{0.130}   & {\underline{0.227}}     & 0.141          & 0.233             & 0.149           & 0.247          & 0.148          & 0.242          & 0.152            & 0.247           & {\underline{0.140}}       & \textbf{0.220}    & 0.162               & 0.253              & 0.184             & 0.289            & 0.147              & 0.240           & 0.153            & 0.249           \\
\multicolumn{1}{c|}{}                             & 336 & \textbf{0.150}   & \textbf{0.241}  & {\underline{0.158}}    & {\underline{0.250}}       & 0.161           & 0.259          & 0.164          & 0.258          & 0.167            & 0.264           & 0.161             & 0.255             & 0.178               & 0.269              & 0.198             & 0.300            & 0.163              & 0.259           & 0.169            & 0.267           \\
\multicolumn{1}{c|}{}                             & 720 & \textbf{0.189}   & \textbf{0.280}  & {\underline{0.190}}    & {\underline{0.281}}       & 0.197           & 0.297          & 0.195          & 0.289          & 0.205            & 0.295           & 0.194             & 0.287             & 0.225               & 0.317              & 0.220             & 0.320            & 0.197              & 0.290           & 0.203            & 0.301           \\ \cline{2-22} 
\multicolumn{1}{c|}{}                             & avg & \textbf{0.149}   & \textbf{0.241}  & {\underline{0.154}}    & {\underline{0.245}}       & 0.160           & 0.259          & 0.159          & 0.254          & 0.165            & 0.260           & 0.156             & 0.247             & 0.178               & 0.270              & 0.193             & 0.295            & 0.159              & 0.253           & 0.166            & 0.264           \\ \hline
\multicolumn{1}{c|}{\multirow{5}{*}{Exchange}}    & 96  & \textbf{0.081}   & \textbf{0.197}  & {\underline{0.083}}    & {\underline{0.199}}       & 0.091           & 0.209          & {\underline{0.083}}    & 0.201          & 0.084            & 0.203           & 0.096             & 0.219             & 0.108               & 0.239              & 0.107             & 0.234            & 0.086              & 0.208           & 0.084            & 0.209           \\
\multicolumn{1}{c|}{}                             & 192 & 0.178            & {\underline{0.298}}     & 0.185          & 0.303             & 0.205           & 0.324          & {\underline{0.171}}    & 0.293          & 0.176            & 0.300           & 0.205             & 0.324             & 0.253               & 0.376              & 0.226             & 0.344            & 0.195              & 0.316           & \textbf{0.162}   & \textbf{0.296}  \\
\multicolumn{1}{c|}{}                             & 336 & 0.343            & 0.421           & 0.345          & 0.422             & 0.353           & 0.431          & \textbf{0.309} & {\underline{0.403}}    & {\underline{0.310}}      & \textbf{0.401}  & 0.378             & 0.456             & 0.390               & 0.471              & 0.367             & 0.448            & 0.342              & 0.425           & 0.333            & 0.441           \\
\multicolumn{1}{c|}{}                             & 720 & {\underline{0.876}}      & {\underline{0.704}}     & 0.877          & 0.706             & 1.115           & 0.801          & \textbf{0.750} & \textbf{0.652} & 0.882            & 0.720           & 1.205             & 0.810             & 1.080               & 0.789              & 0.964             & 0.746            & 0.998              & 0.756           & 0.898            & 0.725           \\ \cline{2-22} 
\multicolumn{1}{c|}{}                             & avg & 0.370            & {\underline{0.405}}     & 0.373          & 0.408             & 0.441           & 0.441          & \textbf{0.328} & \textbf{0.387} & {\underline{0.363}}      & 0.406           & 0.471             & 0.452             & 0.458               & 0.469              & 0.416             & 0.443            & 0.405              & 0.426           & 0.369            & 0.418           \\ \hline
\multicolumn{1}{c|}{\multirow{5}{*}{Illness}}     & 24  & {\underline{1.650}}      & \textbf{0.733}  & 1.777          & 0.788             & 2.076           & 0.956          & 1.877          & 0.800          & 2.728            & 1.114           & 1.693             & 0.872             & 2.743               & 1.150              & 2.317             & 0.934            & \textbf{1.319}     & {\underline{0.754}}     & 2.215            & 1.081           \\
\multicolumn{1}{c|}{}                             & 36  & \textbf{1.173}   & \textbf{0.648}  & {\underline{1.183}}    & {\underline{0.653}}       & 2.183           & 1.008          & 1.508          & 0.725          & 2.669            & 1.092           & 2.002             & 0.942             & 2.887               & 1.183              & 1.972             & 0.920            & 1.430              & 0.834           & 1.963            & 0.963           \\
\multicolumn{1}{c|}{}                             & 48  & \textbf{1.230}   & {\underline{0.694}}     & 1.251          & \textbf{0.686}    & 2.073           & 0.972          & {\underline{1.46}}     & 0.736          & 2.728            & 1.098           & 2.086             & 0.937             & 2.998               & 1.206              & 2.238             & 0.940            & 1.553              & 0.815           & 2.130            & 1.024           \\
\multicolumn{1}{c|}{}                             & 60  & {\underline{1.519}}      & \textbf{0.757}  & 1.538          & {\underline{0.758}}       & 2.058           & 0.974          & 1.663          & 0.781          & 2.883            & 1.126           & 2.102             & 0.946             & 3.160               & 1.234              & 2.027             & 0.928            & \textbf{1.470}     & 0.788           & 2.368            & 1.096           \\ \cline{2-22} 
\multicolumn{1}{c|}{}                             & avg & \textbf{1.393}   & \textbf{0.708}  & {\underline{1.437}}    & {\underline{0.721}}      & 2.098           & 0.978          & 1.597          & 0.731          & 2.752            & 1.108           & 1.971             & 0.924             & 2.947               & 1.193              & 2.139             & 0.931            & 1.443              & 0.798           & 2.169            & 1.041           \\ \hline
\multicolumn{2}{c|}{\textit{1st Count}}                 & \textbf{32}      & \textbf{38}     & 0              & 1                 & 0               & 0              & 4              & {\underline{2}}        & 1                & 1               & {\underline{6}}           & {\underline{2}}           & 0                   & 0                  & 0                 & 0                & 3                  & 0               & 1                & 1               \\ \hline
\end{tabular}
}
  \caption{Performance metrics (MSE and MAE) for multivariate TSF. The best results are highlighted in bold and the second best are underlined. The standard deviation of the metrics across all prediction tasks was controlled within 0.003 with each experiment repeated three times.} 
  \label{table1}
\end{table*}

\subsection{Datasets and Baselines}

\begin{table*}[!t]
\centering 
  \scriptsize 
  \renewcommand{\arraystretch}{0.75}
\begin{tabular}{cc|cc|cc|cc|cc|cc|cc}
\hline
\multicolumn{2}{c|}{Models}                       & \multicolumn{2}{c|}{\textbf{interPDN}} & \multicolumn{2}{c|}{4BSP} & \multicolumn{2}{c|}{BSPDP}     & \multicolumn{2}{c|}{IBBPDP}       & \multicolumn{2}{c|}{SBPDP} & \multicolumn{2}{c}{SBSP} \\ \hline
\multicolumn{2}{c|}{Metric}                       & MSE                & MAE               & MSE         & MAE         & MSE            & MAE            & MSE            & MAE            & MSE             & MAE      & MSE      & MAE           \\ \hline
\multicolumn{1}{c|}{\multirow{5}{*}{ETTh1}} & 96  & \textbf{0.338}     & \textbf{0.374}    & 0.349       & 0.380       & 0.341          & {\underline{0.376}}    & {\underline{0.339}}    & {\underline{0.376}}    & 0.344           & 0.379    & 0.355    & 0.382         \\
\multicolumn{1}{c|}{}                       & 192 & \textbf{0.362}     & \textbf{0.391}    & 0.372       & 0.395       & {\underline{0.365}}    & {\underline{0.392}}    & {\underline{0.365}}    & {\underline{0.392}}    & 0.369           & 0.395    & 0.381    & 0.396         \\
\multicolumn{1}{c|}{}                       & 336 & \textbf{0.378}     & \textbf{0.406}    & 0.394       & 0.410       & {\underline{0.381}}    & {\underline{0.408}}    & 0.382          & 0.409          & 0.383           & 0.410    & 0.398    & 0.413         \\
\multicolumn{1}{c|}{}                       & 720 & \textbf{0.432}     & \textbf{0.451}    & 0.449       & {\underline{0.453}} & {\underline{0.436}}    & {\underline{0.453}}    & 0.439          & 0.456          & 0.445           & 0.459    & 0.446    & 0.456         \\ \cline{2-14} 
\multicolumn{1}{c|}{}                       & avg & \textbf{0.378}     & \textbf{0.406}    & 0.392       & 0.411       & {\underline{0.381}}    & {\underline{0.407}}    & {\underline{0.381}}    & 0.408          & 0.385           & 0.411    & 0.395    & 0.412         \\ \hline
\multicolumn{1}{c|}{\multirow{5}{*}{ETTh2}} & 96  & \textbf{0.223}     & \textbf{0.297}    & 0.227       & 0.301       & 0.225          & {\underline{0.299}}    & {\underline{0.224}}    & \textbf{0.297} & 0.226           & 0.300    & 0.229    & {\underline{0.299}}   \\
\multicolumn{1}{c|}{}                       & 192 & \textbf{0.267}     & \textbf{0.328}    & 0.272       & 0.333       & {\underline{0.268}}    & {\underline{0.329}}    & {\underline{0.268}}    & 0.330          & 0.272           & 0.333    & 0.276    & 0.334         \\
\multicolumn{1}{c|}{}                       & 336 & \textbf{0.305}     & \textbf{0.360}    & 0.310       & 0.364       & {\underline{0.306}}    & {\underline{0.361}}    & {\underline{0.306}}    & \textbf{0.360} & 0.308           & 0.363    & 0.315    & 0.365         \\
\multicolumn{1}{c|}{}                       & 720 & \textbf{0.377}     & \textbf{0.416}    & 0.380       & {\underline{0.417}} & {\underline{0.378}}    & {\underline{0.417}}    & {\underline{0.378}}    & {\underline{0.417}}    & 0.383           & 0.421    & 0.382    & 0.418         \\ \cline{2-14} 
\multicolumn{1}{c|}{}                       & avg & \textbf{0.293}     & \textbf{0.350}    & 0.298       & 0.355       & {\underline{0.294}}    & 0.352          & {\underline{0.294}}    & {\underline{0.351}}    & 0.297           & 0.354    & 0.301    & 0.354         \\ \hline
\multicolumn{1}{c|}{\multirow{5}{*}{ETTm1}} & 96  & \textbf{0.277}     & \textbf{0.331}    & 0.283       & 0.337       & {\underline{0.278}}    & {\underline{0.332}}    & \textbf{0.277} & \textbf{0.331} & 0.288           & 0.336    & 0.280    & 0.333         \\
\multicolumn{1}{c|}{}                       & 192 & \textbf{0.310}     & \textbf{0.352}    & 0.320       & 0.359       & {\underline{0.313}}    & {\underline{0.353}}    & {\underline{0.313}}    & {\underline{0.353}}    & 0.328           & 0.359    & 0.318    & 0.356         \\
\multicolumn{1}{c|}{}                       & 336 & \textbf{0.348}     & \textbf{0.372}    & 0.356       & 0.381       & {\underline{0.349}}    & {\underline{0.373}}    & 0.350          & 0.375          & 0.352           & 0.378    & 0.357    & 0.380         \\
\multicolumn{1}{c|}{}                       & 720 & \textbf{0.411}     & \textbf{0.404}    & 0.424       & 0.414       & \textbf{0.411} & \textbf{0.404} & {\underline{0.416}}    & {\underline{0.407}}    & 0.420           & 0.410    & 0.422    & 0.414         \\ \cline{2-14} 
\multicolumn{1}{c|}{}                       & avg & \textbf{0.337}     & \textbf{0.365}    & 0.347       & 0.374       & {\underline{0.338}}    & {\underline{0.366}}    & 0.339          & 0.367          & 0.347           & 0.371    & 0.344    & 0.371         \\ \hline
\multicolumn{1}{c|}{\multirow{5}{*}{ETTm2}} & 96  & \textbf{0.149}     & \textbf{0.237}    & {\underline{0.151}} & 0.240       & {\underline{0.151}}    & {\underline{0.239}}    & \textbf{0.149} & \textbf{0.237} & 0.165           & 0.249    & 0.151    & 0.240         \\
\multicolumn{1}{c|}{}                       & 192 & \textbf{0.208}     & \textbf{0.277}    & 0.210       & 0.280       & {\underline{0.209}}    & 0.279          & {\underline{0.209}}    & {\underline{0.278}}    & 0.222           & 0.288    & 0.211    & 0.281         \\
\multicolumn{1}{c|}{}                       & 336 & \textbf{0.263}     & \textbf{0.314}    & {\underline{0.264}} & 0.317       & \textbf{0.263} & {\underline{0.315}}    & \textbf{0.263} & {\underline{0.315}}    & {\underline{0.264}}     & 0.317    & 0.267    & 0.318         \\
\multicolumn{1}{c|}{}                       & 720 & \textbf{0.336}     & \textbf{0.362}    & {\underline{0.337}} & 0.366       & 0.339          & {\underline{0.363}}    & 0.339          & {\underline{0.363}}    & 0.338           & 0.365    & 0.343    & 0.368         \\ \cline{2-14} 
\multicolumn{1}{c|}{}                       & avg & \textbf{0.239}     & \textbf{0.298}    & 0.242       & 0.302       & 0.241          & 0.299          & {\underline{0.240}}    & {\underline{0.298}}    & 0.247           & 0.305    & 0.243    & 0.302         \\ \hline
\end{tabular}
  \caption{Ablation results. We respectively ablate the bi-scale branches as well as the interleaved set branches, and report their metrics. We also compare the performance difference between direct scalar estimation head and probability distribution output head under the single-branch configuration. Moreover, we compare interPDN with the network that merely stacks four branches.} 
  \label{table2}
\end{table*}

We conduct extensive experiments on nine real-world datasets spanning electricity transformer monitoring (ETTh1, ETTh2, ETTm1, ETTm2) \cite{zhou2021informer}, Electricity, Weather, Traffic, Exchange-rate, and Illness.
On the Illness dataset, experiments are conducted with forecast horizons of 24, 36, 48, and 60.
For the other eight datasets, the output sequence lengths are set to 96, 192, 336, and 720.
Performance on the test set is evaluated using the metrics MSE and Mean Absolute Error (MAE).

Nine SOTA TSF models serve as baselines, namely xPatch \cite{stitsyuk2025xpatch}, RAFT \cite{han2025retrieval}, AMD \cite{hu2025adaptive}, MOMENT \cite{goswami2024moment}, TimeMixer \cite{wang2024timemixer}, iTransformer \cite{liu2023itransformer}, TimesNet \cite{wu2022timesnet}, PatchTST \cite{nie2022time}, and DLinear \cite{zeng2023transformers}.
Among them, AMD, RAFT and TimeMixer all embody the concept of multi-scale and multi-branch fusion.
The backbone of xPatch exhibits a similar architecture to the backbone of each branch in interPDN.
To validate whether the lightweight interPDN is capable of outperforming TSF foundation models, we select the SOTA model MOMENT as our baseline.

\subsection{Main Results}

Table \ref{table1} presents the evaluation metrics for multivariate TSF achieved by interPDN and other baseline models following parameter search.
Across the 45 forecasting tasks across nine datasets, interPDN achieves first place in MSE on 32 tasks and delivers the best MAE performance on 38 tasks, as shown at the bottom of Table \ref{table1}.
Quantitatively, interPDN achieves SOTA performance on 71.11\% of tasks for MSE and 84.44\% for MAE metrics.
In the remaining prediction tasks where interPDN does not achieve top performance, it still secures second place in eight tasks based on the MSE and six tasks based on the MAE.

Regarding performance improvement margins, interPDN demonstrates a 1.51\% lower MAE and a 2.44\% lower MSE compared to xPatch.
Compared to Transformer-based iTransformer and PatchTST, interPDN achieves a reduction of 35.15\% and 5.31\% in MSE, along with a reduction of 20.27\% and 7.15\% in MAE, respectively.
If compared with RAFT, AMD, and MOMENT which are all latest SOTA baselines, the average reduction rates in MAE are 13.96\%, 4.54\%, and 15.65\%, respectively.

\subsection{Ablation Study}

We conducted ablation studies across four ETT datasets, respectively evaluating the effectiveness of: Single-Branch Scalar Prediction (SBSP), Single-Branch Probabilistic Distribution Prediction (SBPDP), Interleaved Bi-Branch Probabilistic Distribution Prediction (IBBPDP), Bi-Scale Probabilistic Distribution Prediction(BSPDP), and 4-Branch Scalar Prediction aggregated by averaging (4BSP).
The full ablation results are shown in Table \ref{table2}.

SBSP and SBPDP are both conducted using a single-branch backbone model, adding the scalar estimation head and the per-step discrete distribution output head separately.
For 65\% of the prediction tasks, employing the probabilistic distribution output head and computing the expectation outperforms SBSP which directly outputs scalar.
The standalone probabilistic prediction head offers limited improvements to model performance while, in certain cases, turns to degrade the model predictions.

If the interleaved support sets are employed, IBBP model consistently outperforms both SBSP and SBPDP across all tasks in MSE and MAE.
Based on same support set, when augmenting the single-branch probabilistic prediction head with a coarse temporal scale branch as a constraint, this modification also enables BSPDP to yield significant performance improvements.
The proposed interPDN combining the interleaved support set design and the bi-scale scheme, universally outperforms the above-mentioned two network architectures on all prediction tasks.
Furthermore, to demonstrate that the success of interPDN is not merely due to an increase in parameters, we naively integrate the scalar forecasting results from the four-branch architecture. In fact, on 45\% of the forecasting tasks, 4BSP underperforms even SBSP and significantly underperforms interPDN across all tasks.
In summary, each branch as well as the probabilistic prediction head within interPDN is essential.

\begin{figure}[t]
\centering
\includegraphics[width=0.9\columnwidth]{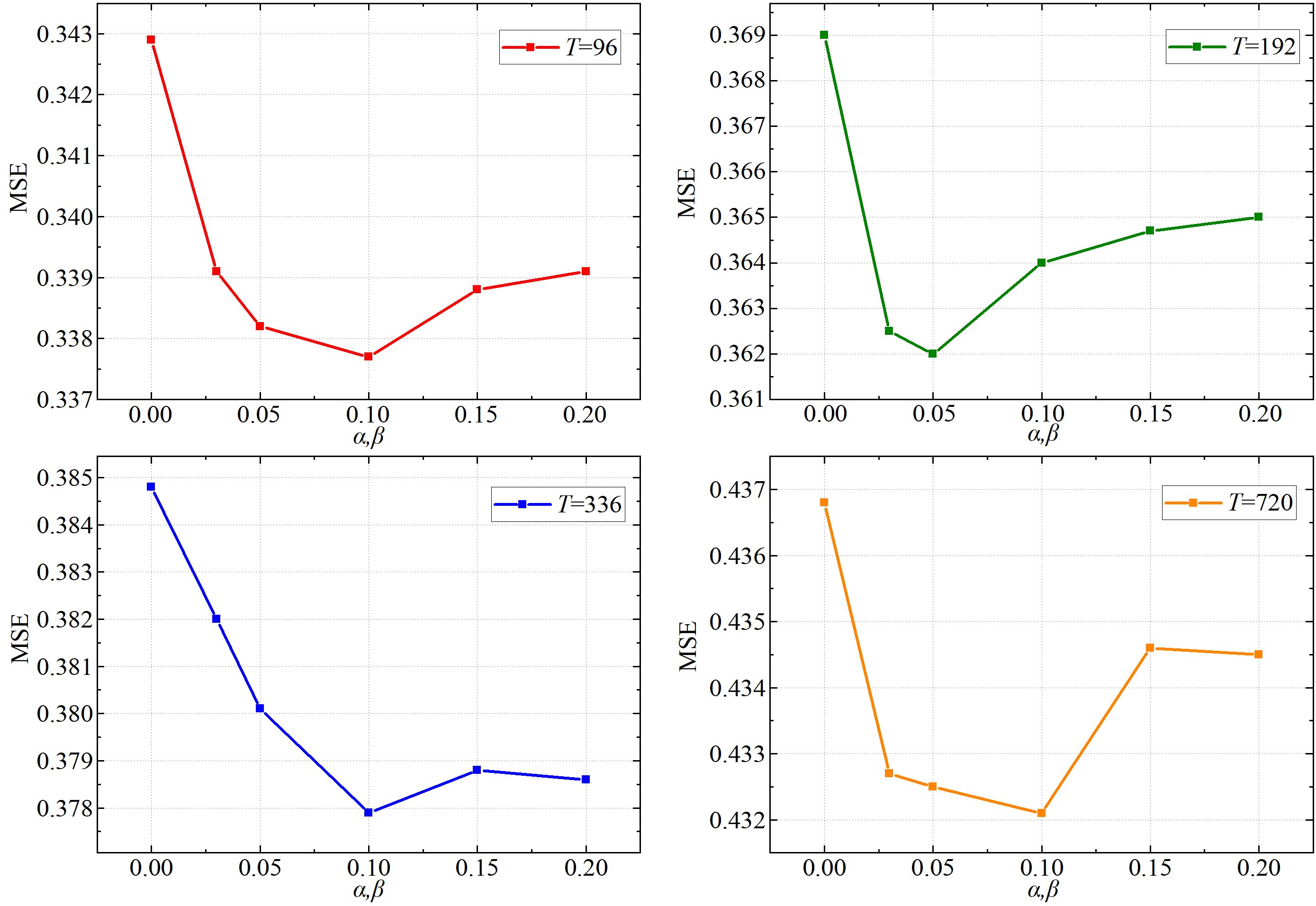} 
\caption{MSE versus $\alpha$ and $\beta$ at four forecast horizons on ETTh1 dataset}
\label{fig3}
\end{figure}

\begin{figure}[t]
\centering
\includegraphics[width=0.9\columnwidth]{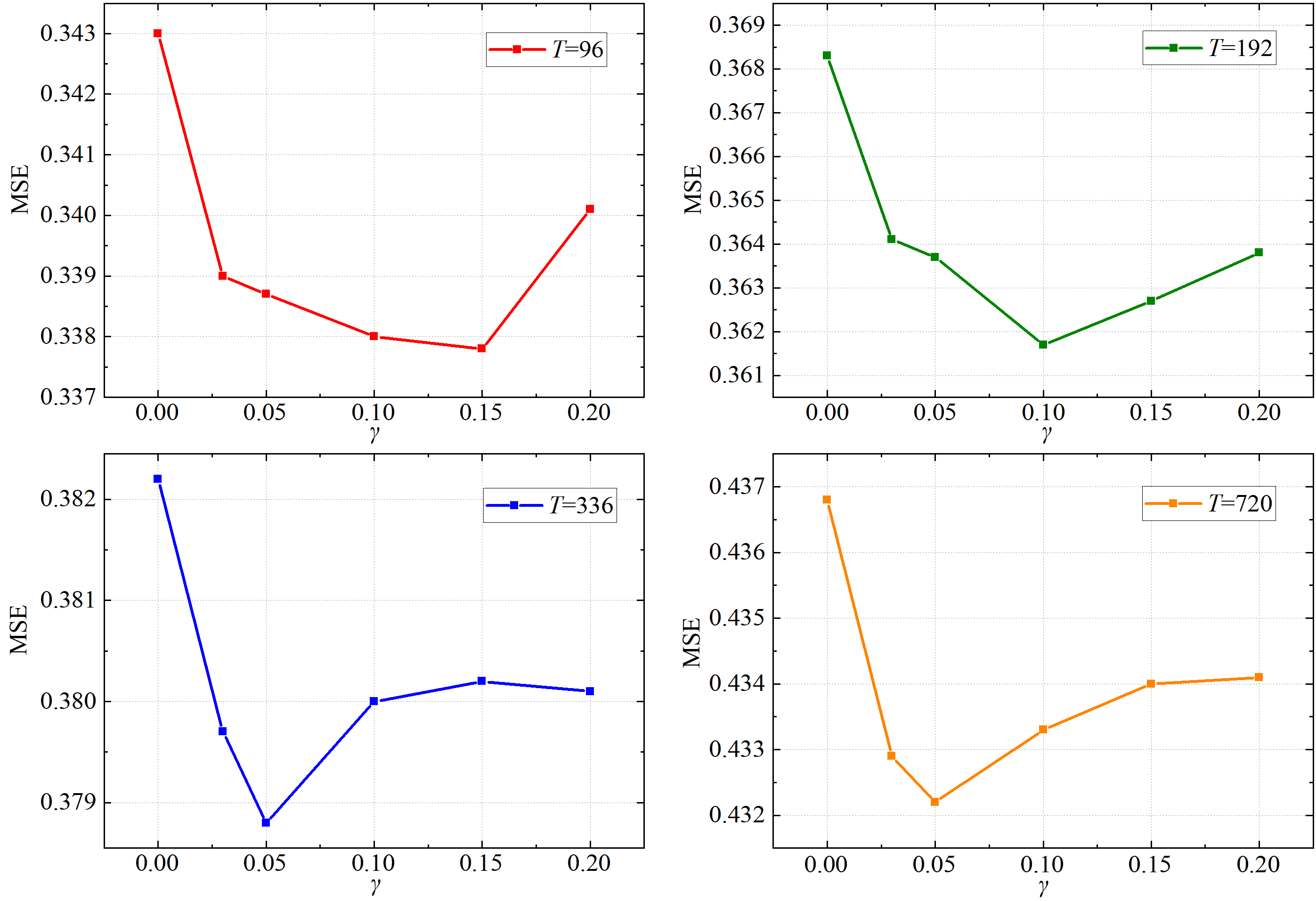} 
\caption{MSE versus $\gamma$ at four forecast horizons on ETTh1}
\label{fig4}
\end{figure}

It is also necessary to verify the effectiveness of constraining the outputs of the interleaved branches using loss functions under the same time scale.
Adjusting the weighting coefficients preceding $L_{f}$ and $L_{c}$ while fixing $\gamma$ to 0.1, we depict the resulting variation in MSE as illustrated in Figure \ref{fig3}.
Since $\alpha$ and $\beta$ are functionally identical consistency losses, we set $\alpha=\beta$ during our hyperparameter search for simplicity.
Across all four TSF tasks on the ETTh1 dataset, MSE is the highest when $\alpha=\beta=0$.
This indicates that without the constraint, the dual branches face difficulties in achieving complementary prediction and effective self-supervision.

Subsequently, while fixing the value of $\alpha$ and $\beta$, we adjust the weighting coefficient $\gamma$ for loss $L_{t}$ to investigate the impact of the inter-scale consistency loss on model performance.
The results in Figure \ref{fig4} exhibit a similar trend to those in Figure \ref{fig3}, with MSE peaking when $\gamma=0$.
It demonstrates that relying solely on fine-scale branch predictions hinders the model’s ability to capture long-term temporal patterns.

\section{Conclusion}

In summary, the proposed interPDN innovatively employs probabilistic distribution prediction and expectation calculation for TSF tasks, fully considering the uncertainty in backbone outputs.
Dual-branch architecture with interleaved support sets is adopted to mitigate quantization errors at adjacent bin boundaries.
A coarse time-scale prediction branch is introduced to enable the model to capture long-range dependencies.
Building on self-supervised principles, two kinds of consistency losses are applied across branches, effectively alleviating single-branch prediction anomalies.
Extensive experiments on multiple real-world benchmarks validate our model's performance, while ablation studies demonstrate the efficacy of probabilistic prediction head, interleaved bi-branch architecture, and inter-scale constraints.
This framework advances TSF by offering a scalable and reliable solution without prior distributional assumptions, thus setting a new direction for TSF in real-world applications ranging from energy to healthcare domains.

\section{Acknowledgments}
We thank the anonymous reviewers for their insightful and constructive comments.
We gratefully acknowledge the support part by the National Natural Science Foundation of China (62376070, 62076195), and part by the Fundamental Research Funds for the Central Universities (AUGA5710011522).

\bibliography{aaai2026}

\newpage
\begin{onecolumn}
\begin{center}
    \LARGE\bfseries Supplementary for Time Series Forecasting via Direct Per-Step Probability Distribution Modeling
\end{center}
\vspace{2em}


\begin{appendices}

The supplementary material begins with a detailed introduction to all datasets used in the main text, the search ranges for hyperparameters, and the final selected parameter configurations.
We also list the computational equipments on which these experimental results are obtained.
A concrete example is provided to illustrate how the interleaved dual-branch structure corrects quantization errors at the bin boundaries corresponding to the support points.
To further demonstrate the ability of the interPDN model in capturing long-term trends, we randomly select several test samples for visualization.
Additional metrics are included to evaluate interPDN's capacity for quantifying uncertainty and its ability to forecast cross-cycle evolutionary patterns.
We replace the backbone models in each branch to demonstrate the generalizability of our approach across multiple baseline architectures.
Experiments with regular support sets are also conducted to validate the effectiveness of the equal-probability support set design.
Computational efficiency comparisons are provided to address potential concerns regarding inference speed and memory usage of the four-branch architecture.


\section{Dataset descriptions}

Details of the nine public datasets used in our experiments are presented in Table \ref{table3}.
The splits into training, validation, and test sets for our model's experiments are identical to those used for all baseline models, as shown in the Dataset Size column of Table \ref{table3}.
The selected datasets include both large-scale, multivariate datasets (e.g., Traffic and Electricity) and lightweight datasets with fewer variables (e.g., ETT and Illness).
Extensive experiments across diverse domains and scales demonstrate our model’s generalization capability for LSTF tasks.

\begin{table}[h]
 \centering 
    \small 
\begin{tabular}{c|ccccc}
\hline
Dataset      & Dim & Series Length         & Dataset Size          & Frequency & Domain         \\ \hline
ETTh1, ETTh2 & 7   & \{96, 192, 336, 720\} & (8545, 2881, 2881)    & 1 hour    & Temperature    \\
ETTm1, ETTm2 & 7   & \{96, 192, 336, 720\} & (34465, 11521, 11521) & 15 min    & Temperature    \\
Weather      & 21  & \{96, 192, 336, 720\} & (36792, 5271, 10540)  & 10 min    & Weather        \\
Traffic      & 862 & \{96, 192, 336, 720\} & (12185, 1757, 3509)   & 1 hour    & Transportation \\
Electriciry          & 321 & \{96, 192, 336, 720\} & (18317, 2633, 5261)   & 1 hour    & Electricity    \\
Exchange     & 8   & \{96, 192, 336, 720\} & (5120, 665, 1422)     & 1 day     & Finance        \\
Illness      & 7   & \{24, 36, 48, 60\}    & (617, 74, 170)        & 1 week    & Healthcare     \\ \hline
\end{tabular}
  \caption{Dataset descriptions. The dataset size is organized in (Train, Validation, Test).} 
  \label{table3}
\end{table}

\section{Hyperparameter Tuning}

The MSE and MAE prediction metrics of our model in Table 1 is obtained through hyperparameter search.
Patch sizes are selected from the range of 6 to 32, with the stride length fixed to half the patch size.
Lookback window lengths are chosen from the range of 96 to 720.
The initial learning rate is tuned between 0.0001 and 0.01.
While the double Sigmoid difference decay is applied exclusively to the Illness dataset, the following decay formula $lr= lr \times 0.9^{(epoch-3)}$ is adopted is for all other datasets.
For the four ETT datasets and the Weather dataset, batch sizes are adjusted within the range of 512 to 2048.
For the remaining datasets, batch sizes are varied between 16 and 256.

Parameters $\alpha$, $\beta$, and $\gamma$ serve as key hyperparameters in our model, governing the weighting of intra-scale consistency loss ($L_f$ and $L_c$) and inter-scale consistency loss ($L_t$) within the final objective function.
Although derived from divergent scales, both $L_f$ and $L_c$ constitute a dual-branch consistency loss operating on the identical interleaved support sets.
To improve the efficiency of parameter search, we therefore constrain $\alpha=\beta$.
The three loss weights are selected from a discrete range of 0 to 0.4.
Given the considerably larger parameter size of the four-branch model, the training epochs are set to exceed 80.
Early stopping would be triggered when the validation loss $L_{total}$ exhibits no reduction over 10 consecutive epochs.
The final hyperparameters selected for all tasks are summarized in Table \ref{table4}.

\begin{table*}[h]
 \centering 
   \small 
   \renewcommand{\arraystretch}{1}
\begin{tabular}{c|c|c|c|c|c|c|c|c|c|c}
\hline
Dataset                   & Pred len & Look back & Initial lr & Batch & $\alpha$    & $\beta$    & $\gamma$    & Patch & Stride & Epochs \\ \hline
\multirow{4}{*}{ETTh1}    & 96       & 512       & 0.0001     & 1024  & 0.05 & 0.05 & 0.1  & 16    & 8      & 100     \\
                          & 192      & 512       & 0.0001     & 1024  & 0.05 & 0.05 & 0.1  & 16    & 8      & 100     \\
                          & 336      & 512       & 0.0001     & 1024  & 0.05 & 0.05 & 0.05 & 16    & 8      & 100     \\
                          & 720      & 512       & 0.0001     & 1024  & 0.1  & 0.1  & 0.1  & 16    & 8      & 100     \\ \hline
\multirow{4}{*}{ETTh2}    & 96       & 512       & 0.0001     & 1024  & 0.02 & 0.02 & 0.1  & 16    & 8      & 100     \\
                          & 192      & 512       & 0.0001     & 1024  & 0.05 & 0.05 & 0.2  & 16    & 8      & 100     \\
                          & 336      & 512       & 0.0001     & 1024  & 0.02 & 0.02 & 0.3  & 16    & 8      & 100     \\
                          & 720      & 512       & 0.0001     & 1024  & 0.02 & 0.02 & 0.4  & 16    & 8      & 100     \\ \hline
\multirow{4}{*}{ETTm1}    & 96       & 512       & 0.0001     & 2048  & 0.3  & 0.3  & 0.2  & 16    & 8      & 150    \\
                          & 192      & 512       & 0.0001     & 2048  & 0.1  & 0.1  & 0.2  & 16    & 8      & 100    \\
                          & 336      & 512       & 0.0001     & 2048  & 0.1  & 0.1  & 0.2  & 16    & 8      & 100    \\
                          & 720      & 512       & 0.0001     & 1024  & 0.1  & 0.1  & 0.1  & 16    & 8      & 100    \\ \hline
\multirow{4}{*}{ETTm2}    & 96       & 512       & 0.0001     & 2048  & 0.1  & 0.1  & 0.1  & 16    & 8      & 100    \\
                          & 192      & 512       & 0.0001     & 2048  & 0.1  & 0.1  & 0.1  & 16    & 8      & 100    \\
                          & 336      & 512       & 0.0001     & 1024  & 0.1  & 0.1  & 0.2  & 16    & 8      & 100    \\
                          & 720      & 512       & 0.0001     & 1024  & 0.1  & 0.1  & 0.2  & 16    & 8      & 100    \\ \hline
\multirow{4}{*}{Weather}  & 96       & 512       & 0.0001     & 512   & 0.1  & 0.1  & 0.2  & 16    & 8      & 100    \\
                          & 192      & 512       & 0.0001     & 1024  & 0.1  & 0.1  & 0.2  & 16    & 8      & 100    \\
                          & 336      & 512       & 0.0001     & 1024  & 0.1  & 0.1  & 0.2  & 16    & 8      & 100    \\
                          & 720      & 512       & 0.0001     & 1024  & 0.05 & 0.05 & 0.2  & 16    & 8      & 100    \\ \hline
\multirow{4}{*}{Traffic}  & 96       & 720       & 0.005      & 64    & 0.1  & 0.1  & 0.2  & 24    & 12     & 120    \\
                          & 192      & 720       & 0.005      & 64    & 0.1  & 0.1  & 0.2  & 24    & 12     & 120    \\
                          & 336      & 720       & 0.005      & 64    & 0.05 & 0.05 & 0.2  & 24    & 12     & 120    \\
                          & 720      & 720       & 0.005      & 64    & 0.05 & 0.05 & 0.3  & 24    & 12     & 120    \\ \hline
\multirow{4}{*}{ECL}      & 96       & 720       & 0.001      & 128   & 0.1  & 0.1  & 0.2  & 16    & 8      & 120    \\
                          & 192      & 720       & 0.001      & 128   & 0.1  & 0.1  & 0.2  & 16    & 8      & 120    \\
                          & 336      & 720       & 0.001      & 128   & 0.1  & 0.1  & 0.2  & 16    & 8      & 120    \\
                          & 720      & 720       & 0.001      & 128   & 0.1  & 0.1  & 0.2  & 16    & 8      & 120    \\ \hline
\multirow{4}{*}{Exchange} & 96       & 96        & 0.0001     & 32    & 0.1  & 0.1  & 0.3  & 16    & 8      & 100    \\
                          & 192      & 96        & 0.0001     & 32    & 0.1  & 0.1  & 0.3  & 16    & 8      & 100    \\
                          & 336      & 96        & 0.0001     & 32    & 0.1  & 0.1  & 0.3  & 16    & 8      & 100    \\
                          & 720      & 96        & 0.0001     & 32    & 0.05 & 0.05 & 0.3  & 16    & 8      & 100    \\ \hline
\multirow{4}{*}{Illness}  & 24       & 36        & 0.01       & 32    & 0.1  & 0.1  & 0.1  & 6     & 3      & 100    \\
                          & 36       & 36        & 0.01       & 32    & 0.1  & 0.1  & 0.3  & 6     & 3      & 100    \\
                          & 48       & 36        & 0.01       & 32    & 0.1  & 0.1  & 0.3  & 6     & 3      & 100    \\
                          & 60       & 36        & 0.01       & 32    & 0.1  & 0.1  & 0.1  & 6     & 3      & 100    \\ \hline
\end{tabular}
  \caption{Comprehensive hyperparameter tuning for LTSF tasks} 
  \label{table4}
\end{table*}

\section{Computing Device Configuration}

Our model is implemented within Python version 3.12 PyTorch version 2.3.0.
Due to their larger number of channels, the Electricity, Weather and Traffic datasets are evaluated on a system with an Intel Xeon Gold 6348 (14 vCPUs) and an NVIDIA A800 GPU (80GB).
Experiments for the remaining six datasets are conducted on a server equipped with an Intel Xeon Gold 6430 (16 vCPUs) and an NVIDIA RTX 4090 GPU (24GB).
The operating system is Ubuntu 22.04.

\section{Effectiveness of Interleaved support sets}

To further illustrate the effect of the interleaved support sets, Figure \ref{fig5} visualizes the discrete probability distributions of dual-branch outputs at a specific time step within the Electricity dataset.
Branch 1 achieves a maximum probability of merely 0.358 on its support set, while the second highest probability reaches 0.333.
This indicates pronounced ambiguity in Branch 1's output decision between support points 0.676 and 0.803, consequently resulting in low prediction confidence.
The 95\% confidence interval of the predicted probability distribution for Branch 2 is 10.35\% narrower than that of Branch 1, and its variance is 24.60\% smaller than that of Branch 1.
Branch 2 outputs the maximum probability of 0.716 at the support point 0.736, while probabilities at all other support points are less than 0.1, indicating high prediction confidence.

The expectation on Branch 1 differs from the true value by 0.022, while the expectation on Branch 2 differs by only 0.007, as shown in the orange dashed box in Figure \ref{fig5}.
After weighted fusion based on maximum prediction confidence, the final output of the dual branches is 0.733, which is 0.003 away from the ground truth.
If relying solely on Branch 1 for prediction, the prediction error would be larger, since the true value lies exactly in the middle of the support points 0.676 and 0.803.
By introducing Branch 2 to output a distribution over another interleaved support sets, it compensates for the quantization error on Branch 1.

\begin{figure*}[h]
\centering
\includegraphics[width=0.85\textwidth]{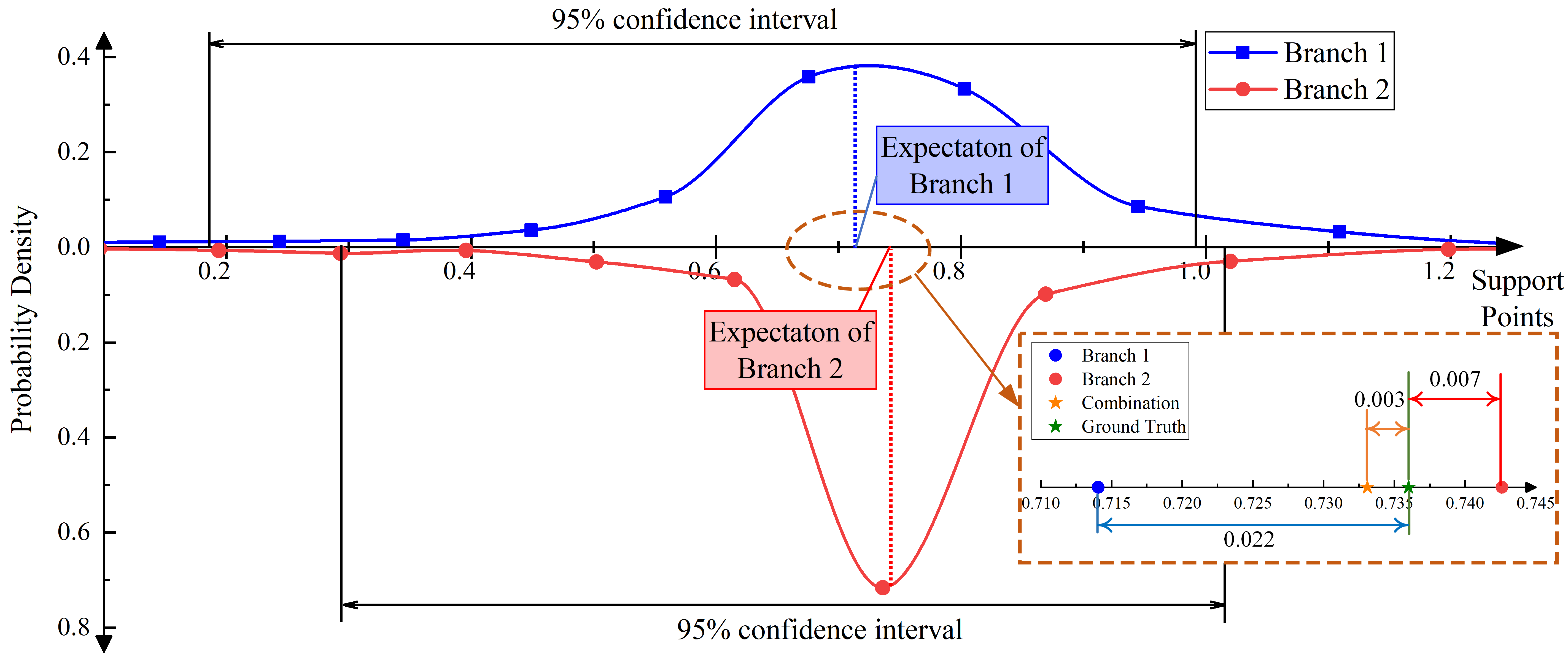}
\caption{Probability distributions predicted by Branch 1 and Branch 2 on the interleaved support sets at a specific time step. An orange dashed box magnifies the interval [0.710, 0.745], showing the expectations predicted by dual branches and the error of the combined prediction relative to ground truth.}
\label{fig5}
\end{figure*}

\section{Long-term Trend Modeling Capability}

To demonstrate that the self-supervised constraint from the coarse-scale branches enables our model to capture long-term trends, Figure \ref{fig6}(a) compares predictions against the ground truth for a test sample in the ETTh1 dataset.
Although TimesNet aggregates cross-period information via 2D convolution, it fits well only on the first 80 time steps and fails to capture long-term trends.
Despite extracting trend components and designing specialized linear module for prediction, xPatch fails to capture the magnitude of upswings.
iTransformer even produces erroneous directional trends contrary to ground truth.
Our model not only achieves the lowest MSE of only 0.112, but also accurately captures the overall upward trend, particularly demonstrating remarkable precision in fitting peak points across multiple cycles.

Subsequently, a representative prediction sample is chosen from the low-frequency Weather dataset. The prediction curves of the aforementioned four models over 336 time steps are plotted in Figure \ref{fig6}(b) for visualization.
Although all models are capable of capturing the periodic oscillations, substantial variations in their accuracy in fitting the amplitude are observed.
Our model's prediction demonstrates the closest alignment with the ground truth, exhibiting significantly lower MSE which is specifically 44.01\%, 38.06\%, and 39.42\% lower than those of TimesNet, xPatch, and iTransformer, respectively.
Overall, with the aid of bi-scale architecture, our model consistently captures long-term trends across datasets of varying frequencies and diverse prediction horizons.

\begin{figure*}[h]
\centering
\includegraphics[width=0.95\textwidth]{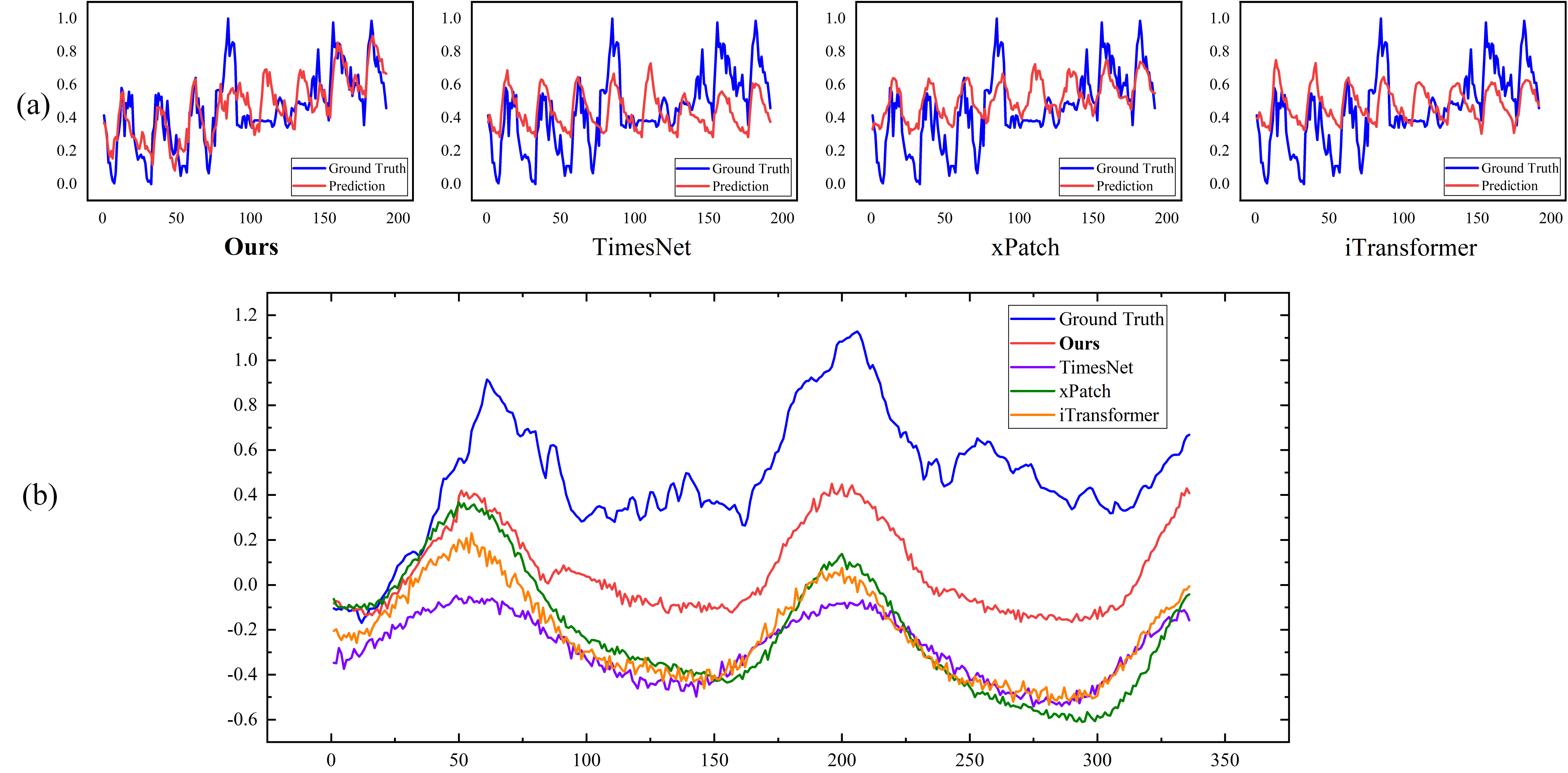}
\caption{(a) Comparison of prediction curves by our model, TimesNet, xPatch, and iTransformer against the ground truth on a test sample ($T=192$) of the ETTh1 dataset. (b) Prediction curves of the four models contrasted with the ground truth on the Weather dataset test sample ($T=336$). The horizontal axis denotes time steps, while the vertical axis represents amplitude.}
\label{fig6}
\end{figure*}

\section{Other Evaluation Metrics}

\begin{figure*}[h]
\centering
\includegraphics[width=0.8\textwidth]{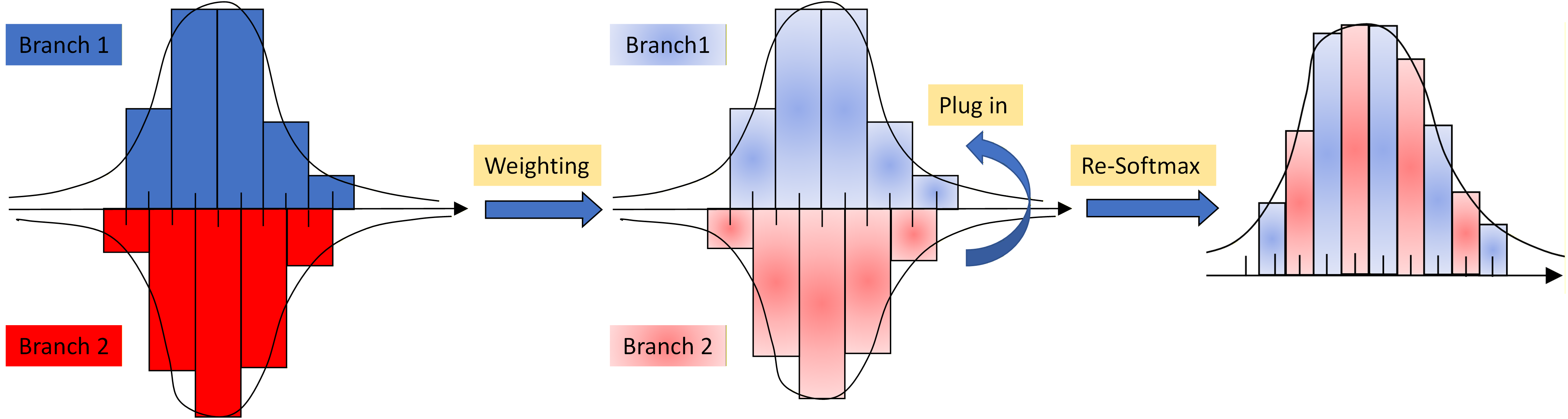}
\caption{Fusion process of dual-branch probability distributions under normal time scales}
\label{com-dual}
\end{figure*}

Since the loss function and evaluation metrics of interPDN are both scalar-based, we supplement the evaluation of the probability distributions of intermediate outputs in this section.
Continuous Ranked Probability Score (CRPS) \cite{gneiting2007strictly} is selected as a probabilistic TSF metric to evaluate interPDN.
Given a predicted distribution with cumulative distribution function $F$ and ground truth $y$, CRPS is defined as:
\begin{equation}
\text{CRPS} = \int_0^1 2\Lambda_\alpha(F^{-1}(\alpha),y)d\alpha,
\end{equation}
where
\begin{equation}
\Lambda_\alpha(q,y) = \max(\alpha(y-q), (1-\alpha)(q-y)).
\end{equation}
$\Lambda_\alpha(q,y)$ represents quantile loss function at quantile level $\alpha \in [0,1]$, and $q$ is the predicted quantile value.

In practice, the CRPS is intractable or computationally expensive to compute, and a normalized metric is required \cite{woo2024moirai}.
A normalized discrete approximation known as the mean weighted sum Quantile Loss (wQL) \cite{park2022learning} is calculated, defined as the average of $K$ quantiles:
\begin{equation}
\text{CRPS}^i \approx \frac{1}{K} \sum_{k=1}^{K} \text{wQL}[\alpha_k]
\end{equation}
\begin{equation}
\text{wQL}[\alpha] = 2 \frac{\sum_{t=1}^{T} \Lambda_{\alpha}(\hat{\mathbf{X}}_{fm,t}^i(\alpha), \mathbf{x}_t^i)}{\sum_{t=1}^{T} |\mathbf{x}_t^i|}
\end{equation}
Here we set $K=9$ quantile levels as well as $\alpha_k \in \{0.1, 0.2, \ldots, 0.9\}$
For the $i$-th channel, $\hat{\mathbf{X}}_{fm,t}^i(\alpha)$ is the predicted mixed probability distribution at time $t$ and quantile level $\alpha$.
$\mathbf{x}_t^i$ represents the ground truth value for the $i$-th channel at time $t$.

As all four branches of interPDN contribute to the final scalar prediction, calculating the CRPS based solely on the probability distribution of any single branch would yield partial results.
Because the two branches at the coarse time scale only serve for self-supervised constraints, we only merge the results $\mathbf{X}_{f,1}^{i}$ and $\mathbf{X}_{f,2}^{i}$ from the two branches at the normal time scale.
As these two probability distributions are defined over interleaved support sets, we first weight them using the fusion weight $\mathbf{w}$ computed by Equation (6) in the paper.
The weighted probability distributions are then interleaved and subsequently processed by a softmax operation to form a fused probability distribution denoted as $\mathbf{X}_{fm}^{i} \in \mathbb{R}^{T \times 2S}$.
As interPDN outputs a discrete probability distribution, it is necessary to first estimate between which two adjacent support points the target quantile lies, and then determine the precise quantile value using linear interpolation.
It is noteworthy that CRPS is employed here solely as an evaluation metric and is not incorporated into the loss function for backpropagation.

In Table \ref{CRPS}, we compare the CRPS of interPDN against four SOTA probabilistic TSF models.
Among them, DiffusionTS \cite{yuan2024diffusion} and TMDM \cite{li2024transformer} are both diffusion model-based approaches for probabilistic time series forecasting.
MOIRAI (base) \cite{woo2024moirai} is a time series foundation model trained on large-scale cross-domain datasets.
Chronos (small) \cite{ansari2024chronos} discretizes time series values into a fixed vocabulary and performs forecasting by leveraging architectures and token generation methods inspired by LLMs.
We fine-tune these pre-trained models using a unified set of hyperparameters (e.g., batch size, lookback window length $L$, patch length $P$, stride length $S_t$, etc.).
On six public datasets, interPDN outperforms specialized probabilistic prediction models on four of them, 
The average CRPS of interPDN is only 3.24\% higher than that of TMDM, while being 39.40\% lower than DiffusionTS, 12.11\% lower than MOIRAI, and 8.29\% lower than Chronos.
These results demonstrate that interPDN retains significant potential even when adapted to the field of probabilistic TSF.

\begin{table*}[h]
 \centering 
   \small 
   \renewcommand{\arraystretch}{1}
\begin{tabular}{c|c|c|c|c|c}
\hline
Models      & \textbf{\begin{tabular}[c]{@{}c@{}}interPDN\\ (Ours)\end{tabular}} & \begin{tabular}[c]{@{}c@{}}DiffusionTS\\ (2024)\end{tabular} & \begin{tabular}[c]{@{}c@{}}TMDM\\ (2024)\end{tabular} & \begin{tabular}[c]{@{}c@{}}MOIRAI\\ (2024)\end{tabular} & \begin{tabular}[c]{@{}c@{}}Chronos\\ (2024)\end{tabular} \\ \hline
ETTh1       & {\underline{0.535}}                                                        & 0.583                                                        & \textbf{0.482}                                        & 0.596                                                   & 0.541                                                    \\
ETTh2       & \textbf{0.417}                                                     & 0.698                                                        & {\underline{0.435}}                                           & 0.488                                                   & 0.505                                                    \\
ETTm1       & {0.547}                                                        & 0.599                                                        & \textbf{0.395}                                        & 0.691                                                   & \underline{0.523}                                                    \\
ETTm2       & \textbf{0.371}                                                     & 0.874                                                        & {\underline{0.381}}                                           & 0.399                                                   & 0.436                                                    \\
Electricity & \textbf{0.397}                                                     & 0.545                                                        & 0.450                                                 & {\underline{0.402}}                                             & 0.481                                                    \\
Traffic     & \textbf{0.507}                                                     & 0.568                                                        & 0.541                                                 & 0.534                                                   & {\underline{0.518}}                                              \\ \hline
\end{tabular}
  \caption{CRPS for probabilistic TSF under a forecast horizon of 192} 
  \label{CRPS}
\end{table*}

Compared to absolute evaluation metrics such as MSE and MAE, the core advantage of the Mean Absolute Scaled Error (MASE) \cite{hyndman2006another} lies in its ability to measure the relative performance of a TSF model against a naive prediction benchmark, and this comparison is scale-invariant.
MASE can be used to determine whether a forecasting model captures more complex evolutionary trend patterns beyond the simplest seasonal patterns present in the data.
Its calculation formula is as follows:
\begin{equation}
\text{MASE} = \frac{1}{C} \sum_{i=1}^{C} \left[ \frac{L - S}{T} \cdot \frac{\sum_{t=1}^{T} \left| \hat{X}^i_t - X^i_t \right|}{\sum_{t=1}^{L-S} \left| X^i_{t+S} - X^i_t \right|} \right]
\end{equation}
where $S$ is a predefined seasonality parameter. For instance, $S=24$ for the hourly-sampled ETTh dataset, and $S=96$ for the ETTm dataset sampled every 15 minutes.
In the code implementation, to prevent exceptionally large MASE values when the denominator approaches zero, an extremely small positive number is added to the denominator.

In Table \ref{MASE}, we supplement the comparison of the MASE of interPDN with that of four other baseline methods.
Since the target window in long-term TSF often contains multiple cycles, it is more prone to anomalous fluctuations and severe distribution shifts compared to short-term forecasting.
As a result, the MASE values in Table \ref{MASE} often exceed 1.
On four ETT datasets, interPDN achieves the lowest MASE across all prediction horizons.
On the most challenging ETTm2 dataset, interPDN reduces MASE by 19.11\% compared to xPatch, and the reduction rates exceed 20\% for the other three baseline models.
This quantitatively demonstrates that interPDN does not simply replicate the periodic patterns from the lookback window, but instead captures the long-term trends evolving across same-phase points of different cycles.

\begin{table*}[h]
 \centering 
   \small 
   \renewcommand{\arraystretch}{1}
\begin{tabular}{cc|c|c|c|c|c}
\hline
\multicolumn{2}{c|}{Models}                       & \textbf{\begin{tabular}[c]{@{}c@{}}interPDN\\ (Ours)\end{tabular}} & \begin{tabular}[c]{@{}c@{}}xPatch\\ (2025)\end{tabular} & \begin{tabular}[c]{@{}c@{}}iTransformer\\ (2024)\end{tabular} & \begin{tabular}[c]{@{}c@{}}PatchTST\\ (2023)\end{tabular} & \begin{tabular}[c]{@{}c@{}}TimesNet\\ (2023)\end{tabular} \\ \hline
\multicolumn{1}{c|}{\multirow{5}{*}{ETTh1}} & 96  & \textbf{0.976}                                                     & {\underline{0.994}}                                             & 1.140                                                         & 1.005                                                     & 1.118                                                     \\
\multicolumn{1}{c|}{}                       & 192 & \textbf{1.026}                                                     & {\underline{1.037}}                                             & 1.235                                                         & 1.127                                                     & 1.181                                                     \\
\multicolumn{1}{c|}{}                       & 336 & \textbf{1.075}                                                     & {\underline{1.089}}                                             & 1.295                                                         & 1.188                                                     & 1.200                                                     \\
\multicolumn{1}{c|}{}                       & 720 & \textbf{1.199}                                                     & 1.211                                                   & 1.359                                                         & {\underline{1.209}}                                               & 1.287                                                     \\ \cline{2-7} 
\multicolumn{1}{c|}{}                       & avg & \textbf{1.069}                                                     & {\underline{1.083}}                                             & 1.257                                                         & 1.132                                                     & 1.197                                                     \\ \hline
\multicolumn{1}{c|}{\multirow{5}{*}{ETTh2}} & 96  & \textbf{1.225}                                                     & {\underline{1.286}}                                             & 2.331                                                         & 1.451                                                     & 1.829                                                     \\
\multicolumn{1}{c|}{}                       & 192 & \textbf{1.379}                                                     & {\underline{1.427}}                                             & 2.246                                                         & 1.499                                                     & 1.938                                                     \\
\multicolumn{1}{c|}{}                       & 336 & \textbf{1.578}                                                     & 1.636                                                   & 2.784                                                         & {\underline{1.577}}                                               & 2.138                                                     \\
\multicolumn{1}{c|}{}                       & 720 & \textbf{1.880}                                                     & 2.018                                                   & 3.107                                                         & {\underline{1.917}}                                               & 2.653                                                     \\ \cline{2-7} 
\multicolumn{1}{c|}{}                       & avg & \textbf{1.516}                                                     & {\underline{1.592}}                                             & 2.617                                                         & 1.611                                                     & 2.140                                                     \\ \hline
\multicolumn{1}{c|}{\multirow{5}{*}{ETTm1}} & 96  & \textbf{0.889}                                                     & {\underline{0.927}}                                             & 1.024                                                         & 1.216                                                     & 1.444                                                     \\
\multicolumn{1}{c|}{}                       & 192 & \textbf{0.957}                                                     & {\underline{1.010}}                                             & 1.156                                                         & 1.277                                                     & 1.079                                                     \\
\multicolumn{1}{c|}{}                       & 336 & \textbf{1.024}                                                     & 1.094                                                   & {\underline{1.087}}                                                   & 1.284                                                     & 1.147                                                     \\
\multicolumn{1}{c|}{}                       & 720 & \textbf{1.117}                                                     & 1.208                                                   & {\underline{1.175}}                                                   & 1.303                                                     & 1.290                                                     \\ \cline{2-7} 
\multicolumn{1}{c|}{}                       & avg & \textbf{0.997}                                                     & {\underline{1.060}}                                             & 1.111                                                         & 1.270                                                     & 1.240                                                     \\ \hline
\multicolumn{1}{c|}{\multirow{5}{*}{ETTm2}} & 96  & \textbf{2.006}                                                     & 3.061                                                   & 3.145                                                         & {\underline{2.956}}                                               & 3.256                                                     \\
\multicolumn{1}{c|}{}                       & 192 & \textbf{2.055}                                                     & 3.089                                                   & 3.123                                                         & {\underline{3.077}}                                               & 3.319                                                     \\
\multicolumn{1}{c|}{}                       & 336 & \textbf{2.446}                                                     & 3.482                                                   & {\underline{2.677}}                                                   & 3.767                                                     & 3.510                                                     \\
\multicolumn{1}{c|}{}                       & 720 & \textbf{2.295}                                                     & 4.022                                                   & 4.007                                                         & {\underline{3.879}}                                               & 4.122                                                     \\ \cline{2-7} 
\multicolumn{1}{c|}{}                       & avg & \textbf{2.201}                                                     & 3.414                                                   & {\underline{3.238}}                                                   & 3.420                                                     & 3.552                                                     \\ \hline
\end{tabular}
  \caption{MASE across the four ETT datasets} 
  \label{MASE}
\end{table*}

\section{Base model replacement experiment}

\begin{table*}[h]
 \centering 
   \small 
   \renewcommand{\arraystretch}{1}
\begin{tabular}{cc|cc|cc|c|c}
\hline
\multicolumn{2}{c|}{Models}                        & \multicolumn{2}{c|}{DLinear}                                & \multicolumn{2}{c|}{interPDN+DLinear} &                                                                                     &                                                                                     \\ \cline{1-6}
\multicolumn{2}{c|}{Metric}                        & MSE                          & MAE                          & MSE               & MAE               & \multirow{-2}{*}{\begin{tabular}[c]{@{}c@{}}MSE Reduction\\ Rate (\%)\end{tabular}} & \multirow{-2}{*}{\begin{tabular}[c]{@{}c@{}}MAE Reduction\\ Rate (\%)\end{tabular}} \\ \hline
\multicolumn{1}{c|}{}                        & 96  & {0.384} & {0.405} & 0.366             & 0.397             & 4.69                                                                                & 1.98                                                                                \\
\multicolumn{1}{c|}{}                        & 192 & {0.405} & {0.413} & 0.400             & 0.418             & 1.23                                                                                & -1.21                                                                               \\
\multicolumn{1}{c|}{}                        & 336 & {0.447} & {0.448} & 0.422             & 0.433             & 5.59                                                                                & 3.35                                                                                \\
\multicolumn{1}{c|}{}                        & 720 & {0.471} & {0.487} & 0.463             & 0.473             & 1.70                                                                                & 2.87                                                                                \\ \cline{2-8} 
\multicolumn{1}{c|}{\multirow{-5}{*}{ETTh1}} & Avg & {0.427} & {0.438} & 0.413             & 0.430             & 3.28                                                                                & 1.83                                                                                \\ \hline
\multicolumn{1}{c|}{}                        & 96  & 0.289                        & 0.353                        & 0.279             & 0.340             & 3.46                                                                                & 3.68                                                                                \\
\multicolumn{1}{c|}{}                        & 192 & 0.383                        & 0.418                        & 0.352             & 0.386             & 8.09                                                                                & 7.66                                                                                \\
\multicolumn{1}{c|}{}                        & 336 & 0.448                        & 0.465                        & 0.371             & 0.408             & 17.19                                                                               & 12.26                                                                               \\
\multicolumn{1}{c|}{}                        & 720 & 0.605                        & 0.551                        & 0.423             & 0.450             & 30.08                                                                               & 18.33                                                                               \\ \cline{2-8} 
\multicolumn{1}{c|}{\multirow{-5}{*}{ETTh2}} & avg & 0.431                        & 0.447                        & 0.356             & 0.396             & 17.39                                                                               & 11.36                                                                               \\ \hline
\multicolumn{1}{c|}{}                        & 96  & {0.301} & {0.345} & 0.280             & 0.336             & 6.98                                                                                & 2.61                                                                                \\
\multicolumn{1}{c|}{}                        & 192 & {0.336} & {0.366} & 0.321             & 0.362             & 4.46                                                                                & 1.09                                                                                \\
\multicolumn{1}{c|}{}                        & 336 & {0.372} & {0.389} & 0.357             & 0.384             & 4.03                                                                                & 1.29                                                                                \\
\multicolumn{1}{c|}{}                        & 720 & {0.427} & {0.423} & 0.418             & 0.418             & 2.11                                                                                & 1.18                                                                                \\ \cline{2-8} 
\multicolumn{1}{c|}{\multirow{-5}{*}{ETTm1}} & avg & {0.359} & {0.381} & 0.344             & 0.375             & 4.18                                                                                & 1.51                                                                                \\ \hline
\multicolumn{1}{c|}{}                        & 96  & {0.172} & {0.267} & 0.171             & 0.255             & 0.58                                                                                & 4.49                                                                                \\
\multicolumn{1}{c|}{}                        & 192 & {0.237} & {0.314} & 0.224             & 0.294             & 5.49                                                                                & 6.37                                                                                \\
\multicolumn{1}{c|}{}                        & 336 & {0.278} & {0.338} & 0.279             & 0.333             & -0.36                                                                               & 1.48                                                                                \\
\multicolumn{1}{c|}{}                        & 720 & {0.420} & {0.437} & 0.365             & 0.386             & 13.10                                                                               & 11.67                                                                               \\ \cline{2-8} 
\multicolumn{1}{c|}{\multirow{-5}{*}{ETTm2}} & avg & 0.277                        & 0.339                        & 0.260             & 0.317             & 6.14                                                                                & 6.49                                                                                \\ \hline
\end{tabular}
  \caption{The performance improvement of DLinear after being adapted with the interPDN framework} 
  \label{DLinear}
\end{table*}

To validate the general applicability of the interPDN method, we replace the base model with DLinear in this section.
DLinear also employs season-trend decomposition, but its backbone is simpler, applying only a single linear layer to the seasonal and trend components, respectively.
Following the architecture of interPDN, we extend DLinear to a four-branch structure.
In each branch, the outputs of the seasonal and trend heads are concatenated and fed into the probabilistic generation module.
All remaining modules are retained, including: the probability distribution prediction head, the interleaved support sets, the combine module, the consistency loss on the interleaved sets, the dual-branch structure under coarser time scale, and the inter-scale consistency loss.

We denote the modified model as interPDN+DLinear and compare its performance with the original DLinear model in Table \ref{DLinear}.
interPDN+DLinear outperforms the original model in almost all forecasting tasks, with an average MSE reduction exceeding 3\% across all four ETT datasets.
On the ETTh2 dataset, the reduction in both MSE and MAE exceeds 10\%.
This demonstrates the strong general applicability of our proposed interPDN method when applied to lightweight base models.
However, caution should be exercised when applying interPDN to base models with excessively large parameter sizes or overly complex network architectures.

\section{Supprt Set Defination}
In subsection \ref{IDB}, we elucidate the generation of a non-uniform support set based on the concept of equiprobable intervals.
Figure \ref{sup} illustrates the distinction between these two types of support sets.
To demonstrate the superiority of our proposed non-uniform support set, we compare the errors of interPDN under identical configurations using both types of support sets in Table \ref{table-sup}.
For fairness, we set the boundary of the support set to $B=4$ and the number of points within the support set to $25$ across all experiments.
Except for the forecasting tasks with $L=96$ and $L=720$ on the ETTm1 dataset, the accuracy of interPDN with the non-uniform support set consistently outperforms that with the uniform support set.
The difference in MSE between the two schemes is most pronounced on the ETTh1 dataset, exceeding a statistically significant margin of 1\%.

\begin{figure}[h]
\centering
\includegraphics[width=0.8\columnwidth]{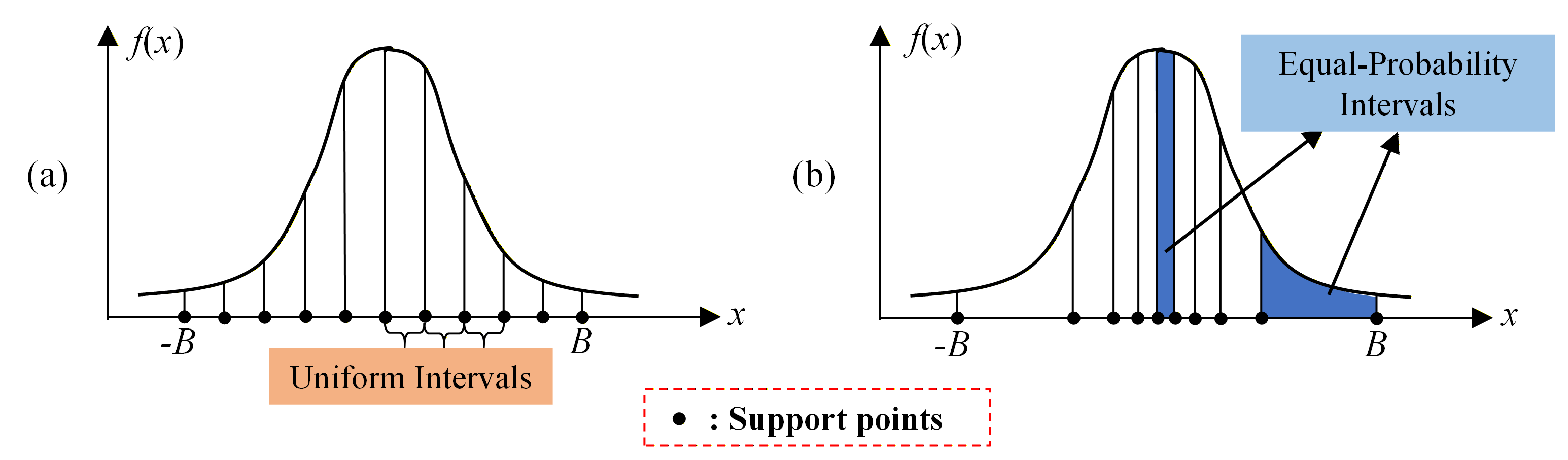} 
\caption{Two methods for partitioning the support set: (a) Extracting support points based on uniform intervals; (b) Partitioning support points using equal-area envelope of the probability density curve.}
\label{sup}
\end{figure}

\begin{table}[h]
 \centering 
   \small 
   \renewcommand{\arraystretch}{1}
\begin{tabular}{cc|cc|cc|c|c}
\hline
\multicolumn{2}{c|}{Models}                       & \multicolumn{2}{c|}{Uniform Set} & \multicolumn{2}{c|}{\textbf{Non-uniform Set}} & \multirow{2}{*}{\begin{tabular}[c]{@{}c@{}}MSE Reduction\\ Rate (\%)\end{tabular}} & \multirow{2}{*}{\begin{tabular}[c]{@{}c@{}}MAE Reduction\\ Rate (\%)\end{tabular}} \\ \cline{1-6}
\multicolumn{2}{c|}{Metric}                       & MSE             & MAE            & MSE               & MAE              &                                                                                    &                                                                                    \\ \hline
\multicolumn{1}{c|}{\multirow{5}{*}{ETTh1}} & 96  & 0.341           & 0.375          & 0.338             & 0.374            & 0.89                                                                               & 0.27                                                                               \\
\multicolumn{1}{c|}{}                       & 192 & 0.368           & 0.392          & 0.362             & 0.391            & 1.66                                                                               & 0.26                                                                               \\
\multicolumn{1}{c|}{}                       & 336 & 0.383           & 0.407          & 0.378             & 0.406            & 1.32                                                                               & 0.25                                                                               \\
\multicolumn{1}{c|}{}                       & 720 & 0.438           & 0.452          & 0.432             & 0.451            & 1.39                                                                               & 0.22                                                                               \\ \cline{2-8} 
\multicolumn{1}{c|}{}                       & avg & 0.383           & 0.407          & 0.378             & 0.406            & 1.32                                                                               & 0.25                                                                               \\ \hline
\multicolumn{1}{c|}{\multirow{5}{*}{ETTh2}} & 96  & 0.224           & 0.298          & 0.223             & 0.297            & 0.45                                                                               & 0.34                                                                               \\
\multicolumn{1}{c|}{}                       & 192 & 0.268           & 0.328          & 0.267             & 0.328            & 0.37                                                                               & 0.00                                                                               \\
\multicolumn{1}{c|}{}                       & 336 & 0.306           & 0.360          & 0.305             & 0.360            & 0.33                                                                               & 0.00                                                                               \\
\multicolumn{1}{c|}{}                       & 720 & 0.380           & 0.418          & 0.377             & 0.416            & 0.80                                                                               & 0.48                                                                               \\ \cline{2-8} 
\multicolumn{1}{c|}{}                       & avg & 0.295           & 0.351          & 0.293             & 0.350            & 0.51                                                                               & 0.21                                                                               \\ \hline
\multicolumn{1}{c|}{\multirow{5}{*}{ETTm1}} & 96  & 0.275           & 0.330          & 0.277             & 0.331            & -0.72                                                                              & -0.30                                                                              \\
\multicolumn{1}{c|}{}                       & 192 & 0.311           & 0.352          & 0.310             & 0.352            & 0.32                                                                               & 0.00                                                                               \\
\multicolumn{1}{c|}{}                       & 336 & 0.348           & 0.372          & 0.348             & 0.372            & 0.00                                                                               & 0.00                                                                               \\
\multicolumn{1}{c|}{}                       & 720 & 0.410           & 0.403          & 0.411             & 0.404            & -0.24                                                                              & -0.25                                                                              \\ \cline{2-8} 
\multicolumn{1}{c|}{}                       & avg & 0.336           & 0.364          & 0.337             & 0.365            & -0.15                                                                              & -0.14                                                                              \\ \hline
\multicolumn{1}{c|}{\multirow{5}{*}{ETTm2}} & 96  & 0.149           & 0.238          & 0.149             & 0.237            & 0.00                                                                               & 0.42                                                                               \\
\multicolumn{1}{c|}{}                       & 192 & 0.209           & 0.278          & 0.208             & 0.277            & 0.48                                                                               & 0.36                                                                               \\
\multicolumn{1}{c|}{}                       & 336 & 0.264           & 0.315          & 0.263             & 0.314            & 0.38                                                                               & 0.32                                                                               \\
\multicolumn{1}{c|}{}                       & 720 & 0.336           & 0.363          & 0.336             & 0.362            & 0.00                                                                               & 0.28                                                                               \\ \cline{2-8} 
\multicolumn{1}{c|}{}                       & avg & 0.240           & 0.299          & 0.239             & 0.298            & 0.21                                                                               & 0.34                                                                               \\ \hline
\end{tabular}
  \caption{Performance improvement of interPDN with non-uniform support sets generated via equal-probability intervals} 
  \label{table-sup}
\end{table}

\section{Computational Efficiency}
Although interPDN incorporates a 4-branch architecture with non-shared parameters and an additional probabilistic generation module, the original backbone remains sufficiently lightweight and structurally simple, comprising only basic MLPs and 1D convolutions.
As a result, it still achieves higher computational efficiency than most baseline models.
Under a batch size of 512 on the same ETTh dataset forecasting task, TimesNet and PatchTST respectively require 10.09 times and 6.78 times more training time per epoch than interPDN, as shown in Table \ref{effi}.
In terms of model scale, their total parameter counts exceed ours by 9.65\% and 6.83\%, while their memory usage is 9.57\% and 11.96\% higher, respectively.
Although interPDN has more parameters than simpler models such as xPatch and DLinear, the marginally higher memory overhead is justified given its significant advantages in metrics such as MSE and MAE.

\begin{table}[h]
 \centering 
   \small 
   \renewcommand{\arraystretch}{1}
\begin{tabular}{c|c|c|c|c}
\hline
Models            & \begin{tabular}[c]{@{}c@{}}Training Time for\\ Each Epoch (s)\end{tabular} & Inference Time (s) & \begin{tabular}[c]{@{}c@{}}Total Number of\\ Parameters\end{tabular} & Memory Usage (MB) \\ \hline
TimesNet          & 12.610                                                                     & 8.030              & 3244007                                                              & 12.37             \\
PatchTST          & 8.418                                                                      & 8.273              & 3160672                                                              & 12.64             \\
xPatch            & 0.775                                                                      & 0.304              & 468656                                                               & 1.79              \\
\textbf{interPDN} & 1.251                                                                      & 0.335              & 2958512                                                              & 11.29             \\ \hline
\end{tabular}
  \caption{Computational efficiency and model scale of interPDN compared to other baselines, when all models are tuned to their best performance via parameter search.} 
  \label{effi}
\end{table}

\end{appendices}
\end{onecolumn}
\end{document}